\documentclass{ieeetj}

\usepackage{cite}
\usepackage{array}
\usepackage[caption=false,font=normalsize,labelfont=sf,textfont=sf]{subfig}
\usepackage{stfloats}
\usepackage{url}
\usepackage{verbatim}
\usepackage{graphicx}
\usepackage{float}
\usepackage{makecell}
\usepackage{bbding}
\usepackage{diagbox}
\usepackage{color}
\usepackage{xcolor}

\usepackage[comma,numbers,square,sort&compress]{natbib}
\usepackage{epstopdf}
\usepackage{amsmath,xparse}
\usepackage{amssymb,amsfonts}
\usepackage{times}
\usepackage{bm}
\usepackage{subfloat}
\usepackage{multirow}
\usepackage{mathtools}
\usepackage{bm}
\usepackage{nomencl}
\makenomenclature
\usepackage{booktabs}
\usepackage{mathrsfs}
\usepackage{amsthm}
\usepackage{lineno}
\usepackage{setspace}
\usepackage{threeparttable}
\usepackage{gensymb}

\usepackage{hyperref}
\usepackage{cleveref}  

\crefname{section}{Sec.}{Secs.}
\crefname{subsection}{Sec.}{Secs.}
\crefname{subsubsection}{Sec.}{Secs.}

\usepackage{twemojis}

\usepackage{algorithmic}
\usepackage{textcomp}
\hypersetup{hidelinks=true}
\usepackage{algorithm}
\def\BibTeX{{\rm B\kern-.05em{\sc i\kern-.025em b}\kern-.08em
    T\kern-.1667em\lower.7ex\hbox{E}\kern-.125emX}}
\AtBeginDocument{\definecolor{tmlcncolor}{cmyk}{0.93,0.59,0.15,0.02}\definecolor{NavyBlue}{RGB}{0,86,125}}

\def\OJlogo{\vspace{-4pt}$<$Society logo(s) and publication title will appear here.$>$}
\def\seclogo{\vspace{10pt}$<$Society logo(s) and publication title will appear here.$>$}

\def\authorrefmark#1{\ensuremath{^{\textbf{#1}}}}

\begin{document}
\receiveddate{XX Month, XXXX}
\reviseddate{XX Month, XXXX}
\accepteddate{XX Month, XXXX}
\publisheddate{XX Month, XXXX}
\currentdate{XX Month, XXXX}
\doiinfo{XXXX.2022.1234567}

\markboth{}{Author {\textit{et al.}}}

\title{\raggedright Multi-cam Multi-map Visual Inertial Localization: \\ System, Validation and Dataset}

\author{Yufei Wei\authorrefmark{1$\dagger$},
Fuzhang Han\authorrefmark{1$\dagger$},
Yanmei Jiao\authorrefmark{2},
Zhuqing Zhang\authorrefmark{1},
Yiyuan Pan\authorrefmark{1}, \\
Wenjun Huang\authorrefmark{1},
Li Tang\authorrefmark{1},
Huan Yin\authorrefmark{3},
Xiaqing Ding\authorrefmark{1},
Chenxiao Hu\authorrefmark{1}, \\
Rong Xiong\authorrefmark{1}, Senior Member, IEEE,
and Yue Wang\authorrefmark{1}, Member, IEEE}

\affil{State Key Laboratory of Industrial Control and Technology, Zhejiang University, Hangzhou, P.R. China}
\affil{School of Information Science and Engineering, Hangzhou Normal University, Hangzhou, P.R. China}
\affil{Hong Kong University of Science and Technology (HKUST), Hong Kong, China}

\corresp{Corresponding author: Yanmei Jiao.}

\begin{abstract}
\textcolor{black}{Robot control loops require causal pose estimates that depend only on past and present measurements. At each timestep, controllers compute commands using the current pose without waiting for future refinements. While traditional visual SLAM systems achieve high accuracy through retrospective loop closures, these corrections arrive after control decisions were already executed, violating causality. Visual-inertial odometry maintains causality but accumulates unbounded drift over time.}
To address the distinct requirements of robot control, we propose a multi-camera multi-map visual-inertial localization system providing real-time, causal pose estimation with bounded localization error through continuous map constraints.
\textcolor{black}{Since standard trajectory metrics evaluate post-processed trajectories,} we analyze the error composition of map-based localization systems and propose a set of evaluation metrics suitable for measuring causal localization performance. To validate our system, we design a multi-camera IMU hardware setup and collect a challenging long-term campus dataset featuring diverse illumination and seasonal conditions.
Experimental results on public benchmarks and on our own collected dataset demonstrate that our system provides significantly higher real-time localization accuracy compared to other methods. To benefit the community, we have made both the system and the dataset open source at \url{https://anonymous.4open.science/r/Multi-cam-Multi-map-VILO-7993}.
\end{abstract}



\maketitle

\section{INTRODUCTION}\label{sec:introduction}

\IEEEPARstart{M}{ap}-based visual-inertial localization (VILO) aims to provide real-time state estimates within a map's reference frame by integrating sensor measurements with information from pre-constructed maps. \textcolor{black}{This capability is fundamental for autonomous robots in field environments, where control loops require immediate position feedback with bounded localization error for safe navigation, precise path following, and reliable return-to-base operations over extended missions.}

\begin{figure}[t]
  \centering
  \setlength{\belowcaptionskip}{-0.3cm}
  \includegraphics[width=0.48\textwidth]{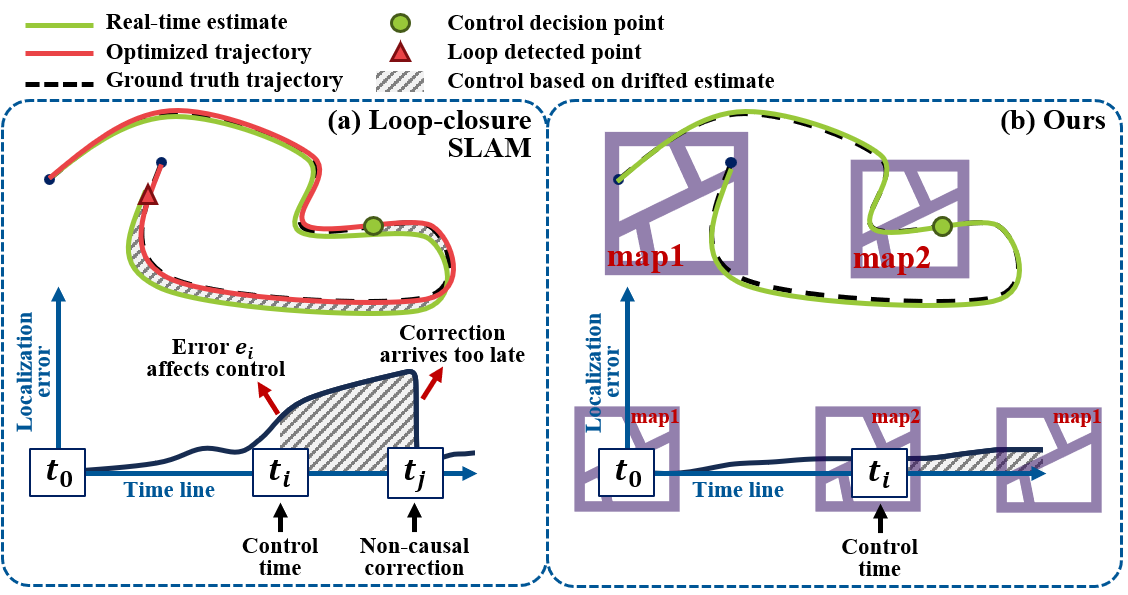}
  \caption{\textcolor{black}{Comparison of localization paradigms for robot control. (a) Traditional SLAM relies on loop closures for drift correction, causing discontinuities and non-causal updates that arrive after control decisions. (b) The proposed multi-map VILO maintains causal, continuous estimates by leveraging pre-built maps for real-time correction, eliminating control-loop incompatibilities.}}
  \label{fig:Causality}
  \vspace{-0.6cm}
\end{figure}

\textcolor{black}{For robot control, VILO systems must provide causal state estimates that depend only on past and present measurements. At each time step, the control loop computes commands using the current pose estimate without waiting for future refinements.
This causality requirement exposes a fundamental tension in existing approaches. Visual-inertial navigation systems (VINS) naturally maintain causality through sequential filtering but inherently accumulate unbounded drift over time~\cite{guoquan1,guoquan2,guoquan3,guoquan4}. Conversely, SLAM systems achieve bounded error through loop closure corrections, but these corrections retroactively modify earlier poses after control decisions were already executed, violating causality, as illustrated in Fig.~\ref{fig:Causality}. This creates three problems for control: drift-affected trajectory execution before correction, discontinuous pose jumps disrupting controllers, and evaluation metrics that mask real-time errors through post-hoc alignment.
This landscape motivates our central question: Can we build a general VILO system that outputs real-time, causal pose estimation that maintains bounded error within the map reference frame and is suitable for control in large-scale, dynamic, and unstructured environments?}
\textcolor{black}{To address this question, we examine the challenges from both system design and performance evaluation perspectives.}

\textcolor{black}{From the system design perspective, achieving causal localization with bounded error requires reconciling three fundamental tensions in system architecture. First, existing map-based localization systems face a scalability-flexibility trade-off in map management. Large-scale monolithic maps are difficult to maintain and update, while existing multi-map systems such as ORB-SLAM3~\cite{orbslam3} and Maplab2.0~\cite{maplab2.0} produce non-causal estimates through batch optimization, with ORB-SLAM3 requiring map overlap and Maplab2.0 requiring additional sensor modalities for non-overlapping map alignment. A practical system needs to support multiple isolated maps without requiring overlap or pre-alignment. Second, long-term deployment in unstructured environments exposes a robustness-efficiency trade-off in continuous localization. Seasonal variations, ongoing construction, and natural weathering can severely degrade feature matching quality, yet the system must maintain real-time performance for control feedback. Third, observation quality and computational efficiency present competing demands. Single-camera systems with limited field-of-view often miss critical structural features, while multi-camera configurations can capture more complete scene information but require efficient integration to maintain real-time operation. These system-level tensions necessitate careful architectural choices that balance causality preservation, error bounding, and computational tractability.}

\textcolor{black}{From the performance evaluation perspective, standard SLAM metrics inadequately reflect the causal estimation quality required for robot control. Metrics such as Absolute Trajectory Error (ATE) and Relative Pose Error (RPE)~\cite{zhang2018tutorial} are typically computed after offline trajectory alignment, which incorporates non-causal post-processing and often neglects errors in the crucial initial transformation to the map frame. This creates a fundamental mismatch: the poses being evaluated are not the poses that controlled the robot. To meaningfully assess localization systems for control applications, evaluation must separately examine each pipeline stage, including map quality, initialization accuracy, and causal trajectory tracking, to isolate error sources and diagnose system behavior under control-relevant constraints.}

\textcolor{black}{In this paper, we propose a multi-camera multi-map VILO system that provides causal pose estimation with bounded localization error for robot control loops. Unlike SLAM systems that retroactively correct drift through batch optimization, our approach integrates map observations directly into each filter update cycle, proactively constraining drift accumulation while preserving causality. By maintaining continuous map constraints rather than periodic corrections, the system ensures bounded error throughout operation. We realize this through four key innovations. First, our filter-based architecture processes observations sequentially, naturally preserving causality. Second, we augment the state vector to track transformations to multiple isolated maps simultaneously, enabling operation across disconnected regions without requiring overlap. Third, we develop IMU-aided minimal solvers that reduce RANSAC sampling requirements from 3-4 correspondences to just 2, maintaining robustness under extreme outlier rates while preserving real-time performance. Fourth, we fuse multi-camera observations directly in the filter update, leveraging expanded field-of-view without additional optimization overhead.}

\textcolor{black}{To properly evaluate our system's control-oriented performance, we propose a comprehensive evaluation framework that analyzes distinct error components in causal, map-based localization. This framework separately assesses underlying map quality, initialization and matching accuracy, and causal trajectory tracking performance, providing insights directly relevant to control applications rather than post-processed reconstruction accuracy.}

Furthermore, to rigorously validate our system's claimed robustness and long-term capabilities, we introduce a new, extensive campus dataset. 
Collected over several months, this dataset is specifically designed to reflect demanding real-world conditions by capturing significant environmental variations such as seasonal changes, diverse illumination, and structural modifications. 
It thus serves as a challenging benchmark for evaluating long-term VILO performance, and we have released it open-source alongside our system to benefit the community.

Overall, the contributions are summarized as follows:
\begin{enumerate}
    \item {We build a multi-cam multi-map VILO system that delivers real-time, causal pose estimation with bounded localization error to the robot position control loop.}
    
    \item {We analyze the error composition in the VILO system from the perspective of robot control and propose to adopt multiple metrics for performance evaluation.}
    \item {We collect a long-term campus dataset featuring multi-session collections conducted over a 9-month period, covering an area of 265,000 square meters with trajectories extending over 55 km.}
    \item {We have open-sourced the proposed VILO system and the collected campus dataset \url{https://anonymous.4open.science/r/Multi-cam-Multi-map-VILO-7993}.}
\end{enumerate}

\textcolor{black}{\textbf{Paper Organization.} The remainder of this paper is organized as follows: Sec.~\ref{sec:related} reviews related work. Sec.~\ref{sec:notation} introduces notation and coordinate systems. Sec.~\ref{sec:system} presents our multi-camera multi-map VILO system, including mapping (Sec.~\ref{sec:mapping}) and localization (Sec.~\ref{sec:localization}) pipelines. Sec.~\ref{sec:metrics} proposes evaluation metrics for causal localization systems. Sec.~\ref{sec:Experiments} provides comprehensive experimental validation, evaluating mapping accuracy (Sec.~\ref{sec:mapping_eval}), matching robustness (Sec.~\ref{sec:matching_eval}), localization performance (Sec.~\ref{sec:localization_eval}), and real-time system comparisons (Sec.~\ref{sec:realtime}). Sec.~\ref{sec:conclusion} concludes the paper.}

\begin{table*}[tp]
\caption{Popular visual SLAM approaches with their supported inputs.}
\centering
\resizebox{0.95\textwidth}{!}{
\begin{threeparttable}
\begin{tabular}{ccccccccccccc}
\hline \hline
\multicolumn{1}{c}{\multirow{2}{*}{Algorithm}} 
& \multicolumn{2}{c}{Camera model support} 
& \multicolumn{4}{c}{Sensor support} 
& \multicolumn{2}{c}{Multi-Map support}
& \multirow{2}{*}{\begin{tabular}[c]{@{}c@{}}\vspace{-0.8ex}External\\ \vspace{-0.8ex}data-source\\ \vspace{-0.8ex}mapping\end{tabular}} 
& \multirow{2}{*}{Open-source} \\
\cmidrule(lr){2-3} \cmidrule(lr){4-7} \cmidrule(lr){8-9}
\multicolumn{1}{c}{}                           & Pinhole               & Fisheye          & IMU & Mono & Stereo & mCam & \begin{tabular}[c]{@{}c@{}}w/ overlap\end{tabular} & \begin{tabular}[c]{@{}c@{}}w/o overlap\end{tabular} &                                                                                                &                              \\
\cline{2-3} \cline{4-7} \cline{8-9}
\hline
RTAB-MAP~\cite{rtab-map2} & $$\checkmark$$  & $$\checkmark$$  &     &      &$$\checkmark$$     &              & $$\checkmark$$   &        &                             & $$\checkmark$$                      \\
ORB-SLAM3~\cite{orbslam3} & $$\checkmark$$  & $$\checkmark$$   &$$\checkmark$$  & $$\checkmark$$  & $$\checkmark$$ &   &$$\checkmark$$ &     &     &$$\checkmark$$   \\
VINS-Fusion~\cite{qin2018online} &$$\checkmark$$     & $$\checkmark$$  &$$\checkmark$$ &      & $$\checkmark$$  &              &    &           & $$\checkmark$$     &$$\checkmark$$  \\
Kimera~\cite{kimera} &$$\checkmark$$     & $$\checkmark$$  &$$\checkmark$$ &      &$$\checkmark$$  &              &   &     &    & $$\checkmark$$    \\
Multi-Col~\cite{multicol} &   & $$\checkmark$$     &     &      &        & $$\checkmark$$    &  &         &  &$$\checkmark$$  \\
VILENS-MC~\cite{VILENS-MC} &$$\checkmark$$  &$$\checkmark$$         & $$\checkmark$$ & $$\checkmark$$  & $$\checkmark$$   & $$\checkmark$$   &  &                           &   &$$\checkmark$$         \\
BAMF-SLAM~\cite{bamf}  &   & $$\checkmark$$   & $$\checkmark$$  &      &        & $$\checkmark$$        &   &                     &                                                                                                &                              \\
MAVIS~\cite{mavis}  & $$\checkmark$$   &$$\checkmark$$   & $$\checkmark$$ & $$\checkmark$$  & $$\checkmark$$  & $$\checkmark$$   &   &        &                    &     \\
Maplab2.0~\cite{maplab2.0} &$$\checkmark$$   &$$\checkmark$$    & $$\checkmark$$ & $$\checkmark$$  & $$\checkmark$$  &    & $$\checkmark$$  &     & $$\checkmark$$     & $$\checkmark$$                 \\
OpenVINS~\cite{openvins} & $$\checkmark$$  &$$\checkmark$$   &$$\checkmark$$ &$$\checkmark$$ & $$\checkmark$$  & $$\checkmark$$   & &       &  & $$\checkmark$$     \\
\hline
\textbf{VILO (ours)}  & $$\checkmark$$  &$$\checkmark$$  & $$\checkmark$$  & $$\checkmark$$  & $$\checkmark$$  &$$\checkmark$$  & $$\checkmark$$ &$$\checkmark$$    & $$\checkmark$$   & $$\checkmark$$ \\
\hline \hline
\end{tabular}
\begin{tablenotes}
    \footnotesize
    \item[1] ``mCam'' means multi-camera; ``mMap'' means multi-map; ``w/'' means with; ``w/o'' means without.
\end{tablenotes}
\end{threeparttable}}
\label{tab:system-compare}
\vspace{-0.4cm}
\end{table*}

\section{RELATED WORK}\label{sec:related}

\textcolor{black}{Visual-inertial localization for robot control requires causal state estimates (depending only on past measurements), flexible multi-region operation, and robustness to long-term appearance changes. Table~\ref{tab:system-compare} summarizes the capabilities of popular visual SLAM systems across key dimensions including camera model support, sensor configurations, and multi-map capabilities. We organize our review around how existing map-based approaches address these requirements, focusing on algorithmic architectures, multi-map capabilities, robust matching techniques, and multi-camera configurations.}


\vspace{-0.2cm}
\subsection{MAP-BASED LOCALIZATION ARCHITECTURES}

Map-based localization systems can be categorized by their algorithmic architecture, which fundamentally determines causality properties and computational characteristics.



\vspace{-0.2cm}
\subsubsection{Optimization-Based Approaches}

ORB-SLAM3~\cite{orbslam3} extends the ORB-SLAM framework with IMU fusion and multi-session capabilities, employing loop closure detection and pose-graph optimization for accurate state estimation. VINS-Fusion~\cite{qin2018online} combines Kalman-filtered initial estimates with pose-graph optimization, supporting monocular, stereo, and fisheye configurations. Kimera~\cite{kimera} decouples visual-inertial odometry from global pose-graph optimization while integrating metric-semantic mapping through parallel modules. RTAB-Map~\cite{rtab-map1} employs memory-managed hierarchical optimization with appearance-based loop closure prioritization for dynamic scene operation. While achieving high precision, these methods produce non-causal estimates, as loop closures retroactively correct earlier poses after control decisions were executed, creating discontinuous pose jumps incompatible with real-time control loops.

\vspace{-0.2cm}
\subsubsection{Filter-Based Approaches}

MSCKF~\cite{msckf} pioneers efficient visual-inertial estimation by marginalizing feature positions from the state, processing observations sequentially to maintain causality. ROVIO~\cite{rovio} integrates patch tracking directly into EKF updates for robustness to fast motion. For large-scale operation, Lynen \textit{et al.}~\cite{lynen2015get} develop descriptor compression techniques, while DuToit \textit{et al.}~\cite{dutoit2017consistent} introduce Cholesky-Schmidt-Kalman filtering with linear memory scaling. OpenVINS~\cite{openvins} provides a modular framework supporting multiple cameras and map-based updates. Geneva \textit{et al.}~\cite{geneva2022map} analyze Schmidt-Kalman filters for map-based localization, demonstrating efficiency in small workspaces and effectiveness of measurement inflation in larger environments. However, these methods typically assume a single map or require overlapping regions for map transitions, limiting multi-region deployment flexibility.

\vspace{-0.6cm}
\textcolor{black}{\subsection{MULTI-MAP OPERATION}}

\textcolor{black}{Large-scale deployment necessitates operation across multiple regions without monolithic map structures. ORB-SLAM3~\cite{orbslam3} supports multiple submaps but requires overlap between maps to enable seamless transitions. Maplab 2.0~\cite{maplab2.0} advances multi-map fusion by aligning non-overlapping trajectories through heterogeneous constraints from LiDAR, semantic objects, and deep features, enabling multi-robot collaboration. However, this approach relies on batch optimization producing non-causal estimates and requires additional sensor modalities beyond visual-inertial observations. No existing filter-based system provides causal localization across multiple isolated maps without overlap requirements using only visual-inertial sensing.}

\vspace{0.2cm}
\subsection{MULTI-CAMERA SYSTEMS AND ROBUST FEATURE MATCHING}

Early multi-camera approaches like MCPTAM~\cite{mcptam} and Multicol-SLAM~\cite{multicol} provide broader spatial coverage but require complex data association unsuitable for resource-constrained platforms. Recent systems have enhanced robustness in challenging environments. VILENS-MC~\cite{VILENS-MC} presents a factor graph optimization-based system using all cameras simultaneously while maintaining a fixed feature budget. BAMF-SLAM~\cite{bamf} processes fisheye imagery to achieve robust state estimation using multiple wide field-of-view cameras. MAVIS~\cite{mavis} introduces an improved IMU pre-integration formulation based on SE2(3) automorphism, enhancing tracking during rapid rotational movement.

\textcolor{black}{However, the effectiveness of multi-camera systems in long-term deployment fundamentally depends on robust feature matching across environmental variations. Seasonal changes, varying illumination, and structural modifications can produce extreme outlier rates, rendering standard RANSAC-based pose estimation with minimal solvers such as P3P~\cite{gao2003complete}, EPnP~\cite{lepetit2009EPnP}, and UPnP~\cite{kneip2014UPnP} computationally prohibitive. The success probability degrades exponentially with outlier rate, requiring thousands of iterations when outliers dominate. While deterministic global methods~\cite{brown2015globally,campbell2017globally,chin2016guaranteed} guarantee convergence, their computational complexity prevents real-time operation. Critically, aggregating observations from multiple cameras increases correspondence count but does not improve inlier rates when environmental appearance has changed, leaving the fundamental convergence limitation unresolved. This gap limits existing multi-camera systems' effectiveness for extended autonomous operation in changing field environments.}


Our work addresses these deployment challenges by developing a lightweight filter-based system capable of consistently fusing multiple isolated maps without overlap requirements, specifically designed for real-time, causal operation in changing environments with limited computational resources. By enhancing feature matching robustness through IMU-aided multi-camera minimal solutions and explicitly modeling pose estimation as a real-time causal system, we provide a framework that delivers position feedback with bounded error suitable for direct integration into robot control loops.



\begin{figure*}[htbp]
  \centering
  \includegraphics[width=0.95\textwidth]{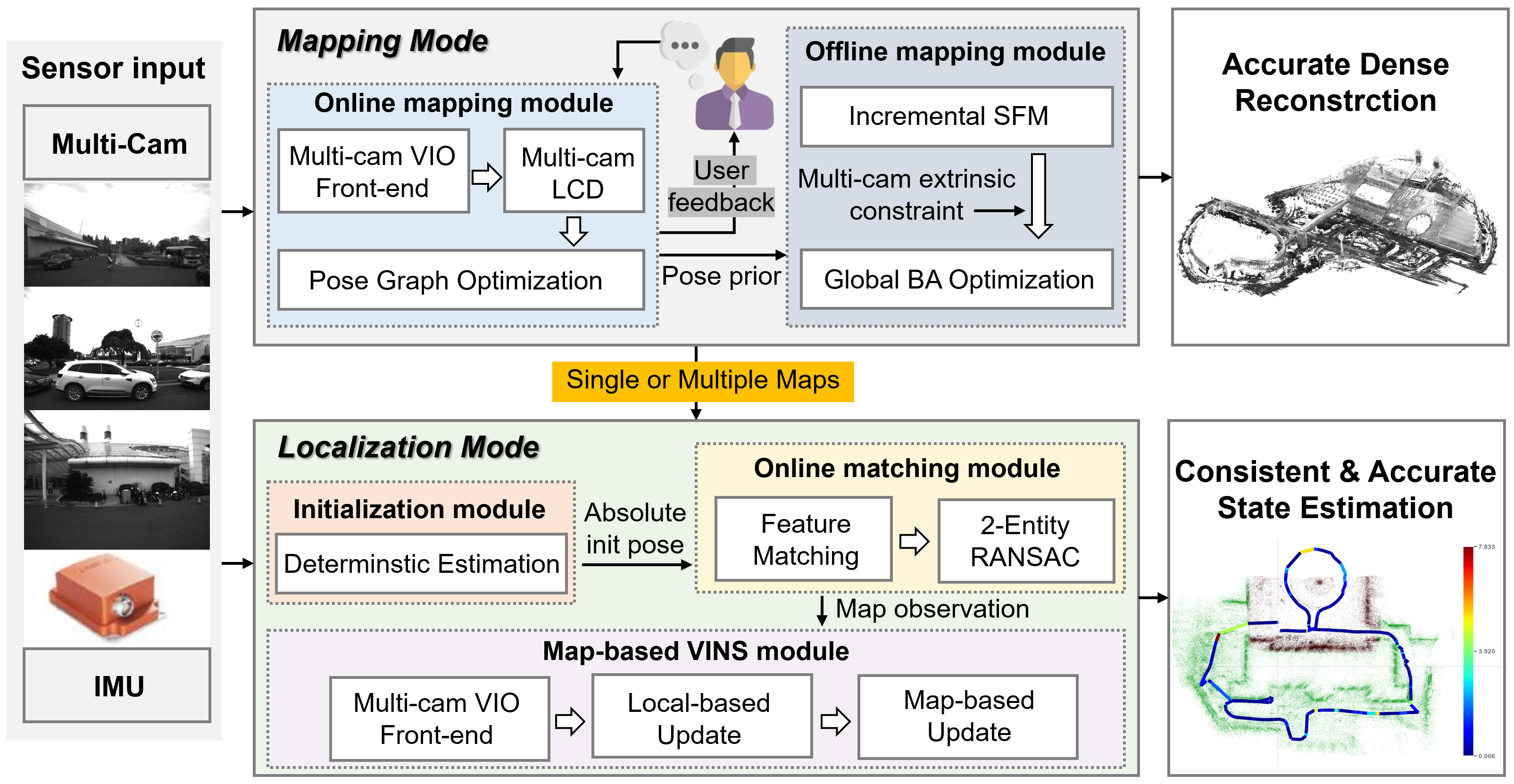}
  \caption{Overview of the proposed VILO system. In mapping mode, the system receives real-time multi-sensor data inputs into the online mapping module for initial map construction, and provides mapping quality feedback to the user, allowing for timely adjustments to the data collection strategy. Then the offline mapping module performs two-stage high-precision map construction and supports dense reconstruction output. In localization mode, the system uses single or multiple isolated maps constructed in mapping mode for map-based robust, accurate and real-time state estimation. Specific details can be found in Sec.~\ref{sec:system}. LCD: Loop closure detection, SFM: structure-from-motion, BA: bundle adjustment.}
  \label{fig:overview}
  \vspace{-0.4cm}
\end{figure*}

\begin{figure}[tp]
  \centering
  \setlength{\belowcaptionskip}{-0.3cm}
  \includegraphics[width=0.45\textwidth]{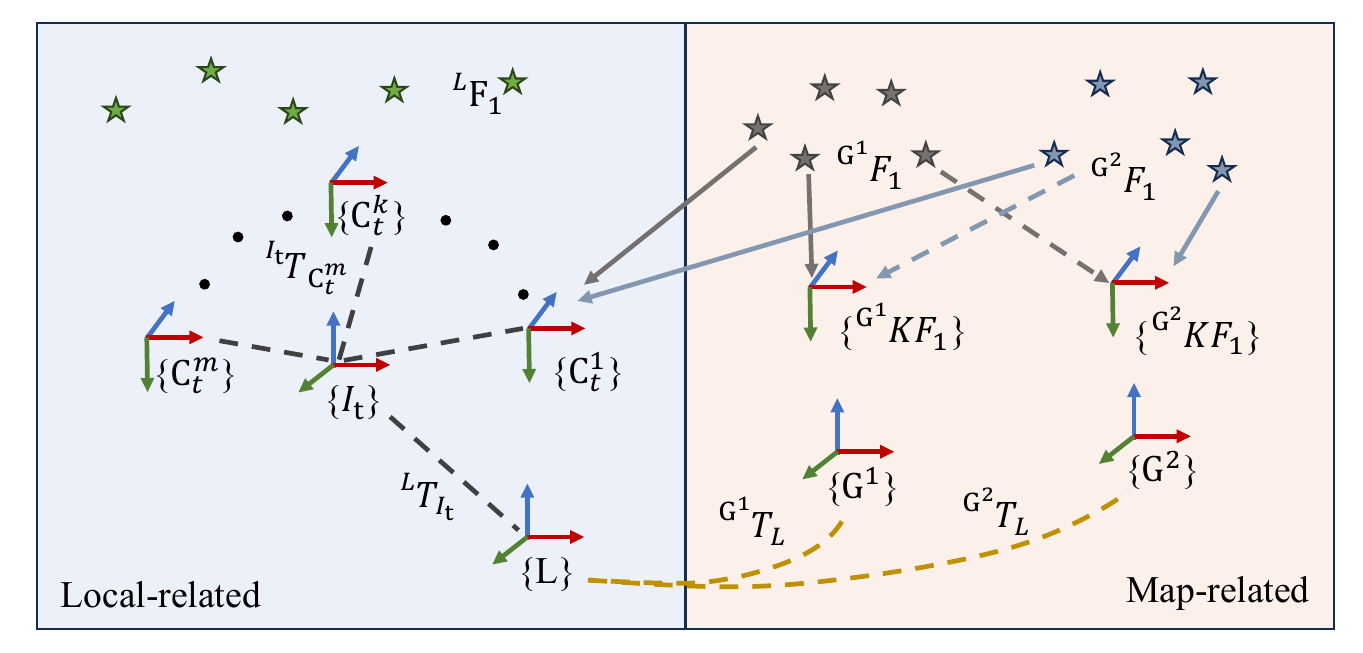}
  \caption{Illustration of each frame. There are two kinds of observations in the proposed system, the local observation (the blue shade part) and the map observation (the pink shade part).}
  \label{fig:frame}
  \vspace{-0.6cm}
\end{figure}

\vspace{-0.4cm}
\section{NOTATION}\label{sec:notation}

\subsection{COORDINATE SYSTEM}

As shown in Table~\ref{tab:reference_frame} and Fig.\ref{fig:frame}, there are five kinds of frames which can be categorized into the local-related frames including $\{L\}$, $\{I_t\}$ and \{$C^{k}_{t}\}$, and map-related frames including $\{G_i\}$ and $\{{}^{G_i}KF_j\}$. For brevity, we consider only two isolated maps on Fig.\ref{fig:frame}. An arrow from the feature point to the frame indicates that the feature is observed at the corresponding time. 

\begin{table}[htbp]
    \centering
    \caption{The definition of involved frames.}
    \label{tab:reference_frame}
    \small
    \renewcommand{\arraystretch}{0.9}
    \begin{tabular}{>{\centering\arraybackslash}m{1.3cm}|>{\raggedright\arraybackslash}p{6.7cm}}
    \hline
    \hline
    \textbf{Frame} & \textbf{Definition} \\
    \hline
    $\{L\}$ & The local inertial frame, which is a fixed frame defined by the initial pose of VIO (cf. $\{L\}$ in Fig.\ref{fig:frame}). \\
    \hline
    $\{I_t\}$ & The IMU (body) frame at timestamp $t$, which is attached to the robot. VIO online estimates the transformation between $\{L\}$ and $\{I_t\}$, i.e., ${}^{L}\mathbf{T}_{I_t}$. \\
    \hline
    $\{C^{k}_{t}\}$ & The $k$-th ($k$ = 1,2,$\cdots$,$m$) camera (image) frame at timestamp $t$. \\
    \hline
    $\{G^{i}\}$ & The $i$-th map frame. All the $i$-th-map-related information (including map keyframes and map features) is based on this frame (cf. $\{G^{1}\}$ and $\{G^{2}\}$ in Fig.\ref{fig:frame}). \\
    \hline
    $\{{}^{G^i}KF_j\}$ & The $j$-th image keyframe of the $i$-th pre-built map, whose pose ${}^{G^{i}}\mathbf{T}_{KF_j}$ is represented in \{$G^{i}$\} (cf. \{${}^{G^1}KF_1$\} and \{${}^{G^2}KF_1$\} in Fig.\ref{fig:frame}). After the $i$-th map is built, the keyframes $\{{}^{G^i}KF_{j}\}$ are fixed. \\
    \hline
    \hline
    \end{tabular}
    \vspace{-0.4cm}
\end{table}

For the transformation between two frames, we use $^{A}\mathbf{T}_B$ to denote the transformation from $\{B\}$ to $\{A\}$ as
\begin{equation}\label{param of time-varying states}
		{^A{\mathbf{T}}_B} \triangleq \begin{bmatrix}
		^A{\mathbf{R}}_B & ^A{\mathbf{p}}_B \vspace {1ex} \\
			\mathbf{0}_{1 \times 3} & 1
		\end{bmatrix}
\end{equation}
where $^A\mathbf{R}_{B}$ is the rotation matrix and $^A\mathbf{p}_B$ represents the translation vector.
Thus, the extrinsic transformation between $\{C^i\}$ and $\{C^{i+1}\}$ is  $^{C^i}\mathbf{T}_{C^{i+1}}$.

\vspace{-0.2cm}
\subsection{SYSTEM STATE}
For the system state definition, we denote the 6 Degree of Freedom (DoF) state, which contains the rotation and translation parts as:
\begin{equation}\label{param2}
	^A\mathbf{x}_{B} = [ \begin{array}{cc}
		^A\mathbf{q}{^{\top}_B} &  ^A\mathbf{p}{^{\top}_B}
	\end{array}]^{\top}
\end{equation}
where $^A\mathbf{q}{_B}$ denotes the JPL quaternion~\cite{trawny2005indirect} and is corresponding to the rotation matrix $^A\mathbf{R}_{B}$ which represents the rotation from the $\{B\}$ to $\{A\}$. $^A\mathbf{p}_B$ represents the translation vector from the $\{B\}$ to $\{A\}$.
Thus, the estimated IMU state is denoted as $^G\mathbf{x}_{I}$, the $1$st to $m$-th camera state is denoted as $^G\mathbf{x}_{C^1}$ to $^G\mathbf{x}_{C^m}$. The extrinsic between IMU and the $k$-th camera is defined as $^I\mathbf{x}_{C^k}$.
Additionally, the position of $i$-th triangulation feature points is denoted as $^G\mathbf{F}_i$. 

\section{MULTI-CAM MULTI-MAP VISUAL INERTIAL LOCALIZATION SYSTEM}\label{sec:system}

\subsection{PROBLEM FORMULATION}\label{sec:problem_formulation}

\textcolor{black}{Map-based visual-inertial localization aims to provide real-time state estimates within a pre-built map's reference frame for autonomous robot navigation. For effective integration into robot control loops, such systems must satisfy three critical requirements simultaneously:}

\textcolor{black}{\textbf{Requirement 1: Causality vs. Drift-bounded.} Control loops require pose estimates that depend only on past measurements without future corrections. VINS methods provide causal estimates but accumulate unbounded drift over time, while SLAM reduces drift through non-causal loop closures that arrive after control decisions, creating pose discontinuities incompatible with real-time control.}

\textcolor{black}{\textbf{Requirement 2: Multi-region operation without overlap.} Robots operating across large-scale environments need flexible map management. Single large maps are difficult to maintain and update, while existing multi-map methods (e.g., ORB-SLAM3, Maplab2.0) require map overlap, limiting operation in disconnected regions.}

\textcolor{black}{\textbf{Requirement 3: Robustness to long-term changes.} Field robots must handle environmental variations across seasons, illumination, and structural modifications. Long-term changes can cause 80\%+ outlier rates in feature matching, overwhelming standard 3-4 point RANSAC methods.}

\textcolor{black}{Our system addresses these requirements through three integrated design choices: (a) a filter-based architecture that fuses map observations in each update cycle, preserving causality; (b) state augmentation to track multiple isolated maps simultaneously without overlap requirements; (c) IMU-aided minimal solvers reducing RANSAC requirements to 2 correspondences. The following subsections detail how these components work together to enable robust localization.}

Generally, a complete VILO system consists of a mapping mode as well as a map-based localization mode. Mapping mode is to obtain the precise pose of each keyframe (e.g. $\{^{G^1}KF_{1}\}$) and position of each triangulated points (e.g. $^{G^1}\mathbf{F}_{1}$) in each map frame (e.g. $\{G^{1}\}$).

As for localization mode, we first need to obtain the transformation from the map frame to the local frame (e.g. $\{^{G^1}\mathbf{T}_L\}$) through initialization module. Then we can combine the results of visual inertial odometry, $^{L}\mathbf{T}_{I_t}$ to get the localization results in map frame. Furthermore, by matching the observations of the current camera with the map keyframes (e.g. \{$C^1$\} and \{$^{G^1}KF_1$\}), we can obtain additional map observations to mitigate the drift of the odometry.

\subsection{SYSTEM OVERVIEW}

\textcolor{black}{Fig.\ref{fig:overview} illustrates our system architecture, designed to meet practical deployment requirements through two operational modes.}

\textcolor{black}{\textbf{Mapping Mode Design Rationale.} Real-world mapping faces a critical challenge: poor data quality during collection (e.g., excessive motion blur) can cause mapping failure, but users often lack expertise to assess quality in real-time. Therefore, we implement a two-stage pipeline:}

\begin{itemize}
    \item \textcolor{black}{\textbf{Online mapping module} (Sec.~\ref{sec:online_map}): Runs lightweight multi-camera SLAM during data collection, providing immediate feedback on mapping quality via velocity monitoring. This allows users to re-collect data when quality issues are detected.}
    
    \item \textcolor{black}{\textbf{Offline mapping module} (Sec.~\ref{sec:offline_map}): After high-quality data collection, performs high-precision reconstruction using two-stage SfM with multi-camera extrinsic constraints. Supports external sensor fusion (LiDAR/GPS) when available for scale consistency.}
\end{itemize}

\textcolor{black}{\textbf{Localization Mode Design Rationale.} Robot control requires real-time causal estimates, motivating our filter-based architecture over optimization-based SLAM. The localization pipeline consists of:}

\begin{itemize}
    \item \textcolor{black}{\textbf{Initialization module} (Sec.~\ref{sec:init}): Deterministic optimal pose estimation ensures reliable startup even under extreme outlier rates ($>$80\%), critical since initialization failure terminates localization capability.}
    
    \item \textcolor{black}{\textbf{Online matching module} (Sec.~\ref{sec:online_match}): IMU-aided 2-point RANSAC provides continuous map observations robust to long-term appearance changes, maintaining real-time performance.}
    
    \item \textcolor{black}{\textbf{Map-based VINS module} (Sec.~\ref{sec:vins}): Filter-based state estimation with augmented state vector tracking multiple isolated map transformations simultaneously, fusing local VIO and map observations to provide causal estimates with bounded error through continuous map constraints.}
    
\end{itemize}

\textcolor{black}{The key innovation is how these components integrate: the deterministic initialization establishes reliable map alignment, the robust matching provides continuous constraints despite environmental changes, and the filter-based architecture fuses everything while preserving causality for control.}

\subsection{MAPPING}\label{sec:mapping}
\textcolor{black}{Our mapping pipeline addresses a critical challenge for practical robot deployment: how to construct accurate maps from multi-camera observations while ensuring data quality and scale consistency. The design follows a two-stage philosophy: real-time quality assessment during data collection, followed by high-precision offline reconstruction. When external sensors (LiDAR/GPS) are available, our modular architecture integrates their measurements to improve scale consistency in large-scale environments.}

\vspace{-0.2cm}
\subsubsection{Online visual mapping module}\label{sec:online_map}
\begin{figure}[tbp]
  \centering
  \includegraphics[width=0.48\textwidth]{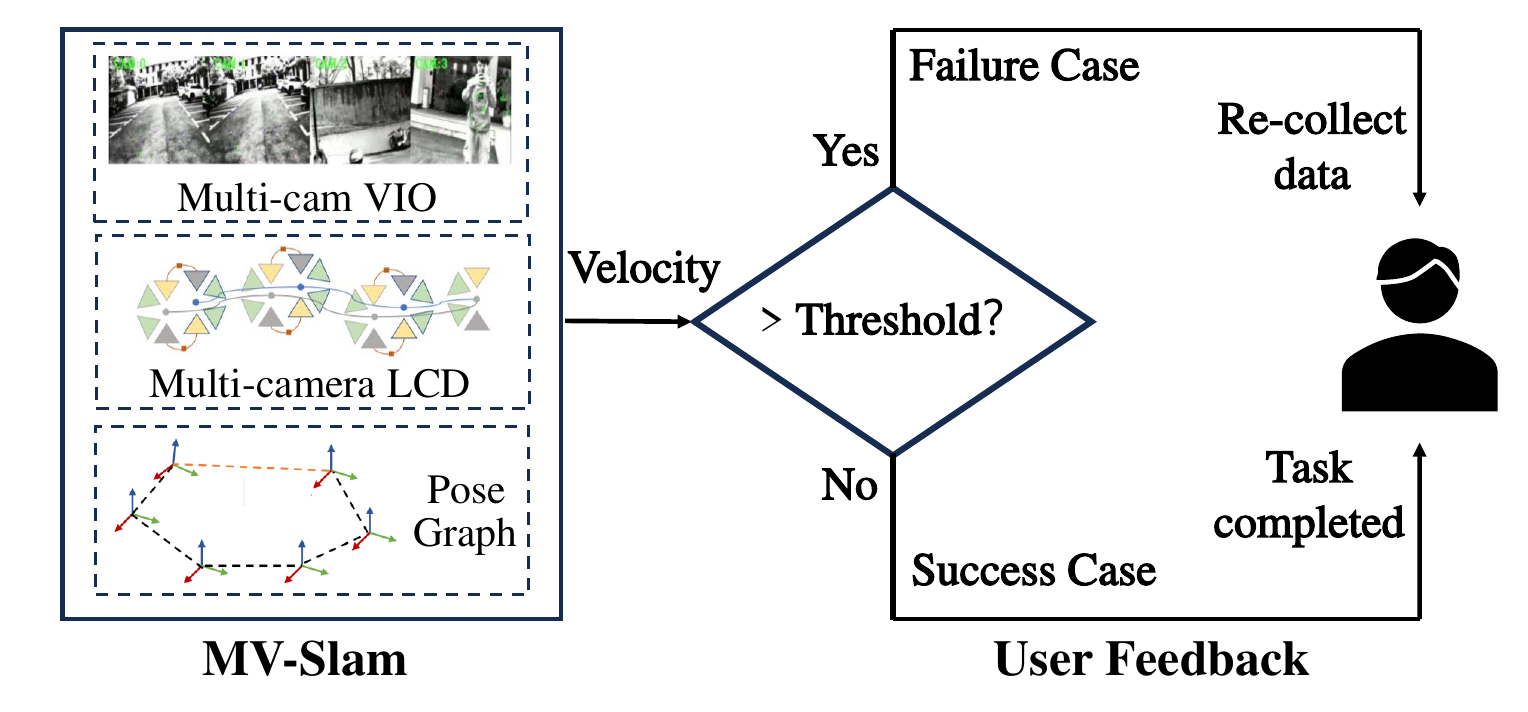}
  \caption{Online mapping process.}
  \label{fig:online_mapping}
  \vspace{-0.4cm}
\end{figure}

\textcolor{black}{The online module (Fig.\ref{fig:online_mapping}) serves a dual purpose: constructing an initial map while providing real-time feedback on data collection quality. This addresses a practical problem in field robotics where users often lack expertise to assess whether collected data will support successful mapping.}

\textcolor{black}{\textbf{Multi-camera SLAM architecture.} We build upon OpenVINS~\cite{openvins} as the VIO front-end, integrating it with loop closure detection and pose graph optimization to address drift accumulation in extended trajectories. The filtering-based VIO enables lightweight real-time processing during data collection, providing immediate pose estimates and quality feedback to users. Loop closures then correct accumulated drift through 4-DoF pose graph optimization. This hybrid architecture ensures both real-time responsiveness for quality monitoring and long-term consistency through optimization.}

\textcolor{black}{A critical adaptation for field environments is multi-camera loop closure detection. Standard monocular configurations often fail in urban/campus scenarios where forward-facing cameras predominantly capture road surfaces with limited structural features. We augment loop detection with observations from lateral cameras, which typically observe building facades and other structured elements, substantially improving closure recall in feature-sparse forward views. The system maintains a sliding window of keyframes from all cameras, queries them against a unified DBoW2~\cite{2012bags} visual vocabulary, and validates geometric consistency through multi-view constraints before triggering loop closures.}

\textcolor{black}{\textbf{Real-time quality feedback.} Excessive robot motion during collection produces two failure modes: motion blur degrading feature detectability, and large inter-frame baselines causing feature matching failures. Both manifest as tracking instability in the VIO front-end. We monitor the estimated velocity magnitude from the VIO state, where sudden increases beyond empirically-determined thresholds indicate either algorithm failure or motion too rapid for reliable reconstruction. Upon detecting these conditions, the system alerts users to either reduce speed or re-collect the problematic segment, ensuring high-quality input for subsequent offline processing.}

\vspace{-0.2cm}
\subsubsection{Offline visual mapping module}\label{sec:offline_map}
\textcolor{black}{The offline module refines the initial online map into high-precision reconstruction suitable for localization. Traditional multi-camera methods fix all cameras in a unified frame, but this propagates extrinsic calibration errors. Our two-stage approach treats extrinsics as soft constraints rather than hard fixes.}

\textcolor{black}{\textbf{STAGE I: VINS-guided incremental SfM.} We employ COLMAP's~\cite{colmap} incremental pipeline with multi-camera VINS trajectory as pose priors. These priors enable spatial matching, which pairs images by spatial proximity rather than sequential order, effectively avoiding mismatches in repetitive urban scenes.}

\textcolor{black}{For scale consistency, we anchor the reconstruction to multi-camera geometry. Specifically, we establish the reference camera's initial pose as the global frame origin, compute all camera poses using calibrated extrinsics, then apply COLMAP's geo-registration to transform the reconstruction into this coherent coordinate system. This prevents arbitrary scale drift common in pure visual methods.}

\textcolor{black}{\textbf{STAGE II: Extrinsic-constrained global optimization.} Using STAGE I output as initialization, we perform global bundle adjustment incorporating multi-camera extrinsic constraints:}
\begin{equation}\label{all cost function}
	E = E_{proj} + E_{extrinsic}
\end{equation}
\textcolor{black}{where $E_{proj}$ is the standard robust reprojection error~\cite{triggs2000bundle}, and $E_{extrinsic}$ penalizes deviations from calibrated inter-camera geometry. Specifically, for each non-reference camera, we compute its estimated relative pose to the reference camera from the current reconstruction, then formulate constraints comparing this estimate against calibrated extrinsics, treating rotation and translation separately. Both terms employ Huber robust cost functions and information matrices encoding measurement uncertainty.}

\textcolor{black}{This soft constraint formulation balances geometric validity with calibration uncertainty. Rigidly enforcing extrinsics as hard constraints injects noise when calibration is imperfect (as demonstrated in Fig.\ref{fig:init_compare}d), while ignoring them discards valuable multi-camera geometric information. The optimized map maintains both reconstruction accuracy and real-world scale consistency, with output compatible with COLMAP's dense reconstruction for downstream applications.}

\vspace{-0.2cm}
\subsubsection{Scale-aware mapping with multi-sensor integration}\label{sec:multi_sensor_map}
\textcolor{black}{Large-scale field deployments expose a fundamental limitation of visual-inertial mapping: scale drift from IMU bias accumulation and poor visual observability in texture-sparse regions. When available, we integrate external sensors to provide absolute scale references.}

\textcolor{black}{\textbf{Modular sensor fusion.} Our architecture processes LiDAR or GPS measurements into 6-DoF pose observations through sensor-specific preprocessing. These external pose observations enhance both offline stages: guiding feature triangulation and initial reconstruction in STAGE I, and contributing additional constraints to global optimization in STAGE II. The modular design maintains operational flexibility, as the system functions with pure visual-inertial input when external sensors are unavailable.}

\textcolor{black}{\textbf{Global frame alignment.} For multi-session consistency, we align the final map to an absolute reference frame via least-squares rigid transformation minimizing pose errors. This enables cross-session localization and collaborative mapping across robot teams.}




As shown in Sec.~\ref{sec:map-exp3}, this multi-sensor approach significantly improves mapping accuracy in large-scale environments and maintains flexibility when certain sensors are unavailable, adapting to various field deployment constraints.

\subsection{LOCALIZATION}\label{sec:localization}
\textcolor{black}{The localization mode aims to achieve robust, accurate, and real-time state estimation in the map reference frame. This section presents a three-stage pipeline that addresses the distinct challenges arising when robots must operate in mapped environments under long-term appearance variations.}

\textcolor{black}{The pipeline begins with an initialization module (Sec.~\ref{sec:init}) that establishes the transformation from the local VIO frame to the map frame with deterministic optimality. Unlike probabilistic RANSAC-based methods that may fail under extreme outlier rates ($>$80\%) caused by seasonal changes or illumination variations, our approach guarantees convergence by decoupling the 4DoF pose estimation problem into 1DoF rotation and 3DoF translation searches. This deterministic guarantee is mission-critical, as initialization failure terminates all subsequent localization capability.}

\textcolor{black}{After successful initialization, the online matching module (Sec.~\ref{sec:online_match}) continuously provides map observations by matching current camera frames against map keyframes. To maintain real-time performance while preserving robustness, we employ IMU-aided minimal solvers that reduce required correspondences from 3-4 points to 2 points, significantly improving RANSAC convergence under challenging conditions. These filtered observations are then fed to the map-based VINS module (Sec.~\ref{sec:vins}), which fuses local VIO estimates with map observations through a filter-based framework that preserves causality. By augmenting the state vector to simultaneously track transformations to multiple isolated maps, the system maintains bounded localization error across disconnected regions without requiring map overlap.}

\textcolor{black}{The key innovation lies in the integration: deterministic initialization ensures reliable startup, robust matching maintains continuous constraints despite appearance variations, and the filter-based architecture fuses everything while preserving the causality required for robot control loops.}

\vspace{-0.2cm}
\subsubsection{Initialization module}\label{sec:init}
\textcolor{black}{The initialization module associates current visual-IMU observations with a pre-built map to compute the camera's pose in the map reference frame, aligning the local VIO origin with the map's coordinate system. This section presents a deterministic pose estimation algorithm designed to handle extreme outlier rates in long-term operation.}


\textcolor{black}{\textbf{Problem Formulation and Requirements.} Successful initialization is mission-critical, as failure prevents all subsequent localization. The key challenge is achieving deterministic convergence under extreme outlier rates ($>$80\%) caused by seasonal appearance changes, illumination variations, and viewpoint shifts. Standard RANSAC-based methods~\cite{choi1997performance} cannot guarantee convergence due to their probabilistic nature, while existing deterministic methods~\cite{brown2015globally,campbell2017globally,chin2016guaranteed} require exhaustive 6DoF search with prohibitive computational complexity when correspondences exceed 100.}

\textcolor{black}{Our approach exploits the vehicle's locally planar motion to reduce dimensionality while maintaining global optimality. Inertial measurements provide observable pitch and roll angles~\cite{li2013high}, enabling gravity-alignment between query and map frames. This reduces the unknown pose from 6DoF to 4DoF, including one yaw angle $\alpha$ and 3D translation $\mathbf{t}$:}
\begin{equation}\label{4dof}
^{G^1}\mathbf{T}_{C^k_1} \triangleq \begin{bmatrix}
\mathbf{R}(\alpha) & \mathbf{t} \vspace {1ex} \\
\mathbf{0}_{1 \times 3} & 1
\end{bmatrix}
\end{equation}
\textcolor{black}{where $\mathbf{R}(\alpha)$ is the rotation matrix parameterized solely by yaw angle $\alpha \in [-\pi, \pi]$.}

\textcolor{black}{\textbf{Two-Point Minimal Solution.} For each 2D-3D correspondence $(\mathbf{u}_i, \mathbf{F}_i)$ between image observation and map point, the transformed 3D point must lie on the projection ray. This collinearity constraint yields:}
\begin{equation}\label{eq.linep}
\begin{cases}
{v}_{i,x}(\mathbf{R}_3 \mathbf{F}_i + t_z) = \mathbf{R}_1 \mathbf{F}_i + t_x \\
{v}_{i,y}(\mathbf{R}_3 \mathbf{F}_i + t_z) = \mathbf{R}_2 \mathbf{F}_i + t_y
\end{cases}
\end{equation}
\textcolor{black}{where $\mathbf{R}(\alpha)\triangleq (\mathbf{R}_1^{\top},\mathbf{R}_2^{\top},\mathbf{R}_3^{\top})^{\top}$ and $\mathbf{v}_i = \mathcal{K}^{-1}[\mathbf{u}_i^{\top}, 1]^{\top}$ is the normalized image coordinate. Two correspondences provide four constraints for four unknowns. By algebraically eliminating translation from these equations, we obtain a translation-invariant measurement constraining only yaw:}
\begin{equation}\label{eq.alpha}
d(\alpha) = d_1 \sin\alpha + d_2 \cos\alpha + d_3 = 0
\end{equation}
\textcolor{black}{where coefficients $d_1, d_2, d_3$ are computed from correspondence geometry. Each correspondence pair yields an analytical yaw solution via (\ref{eq.alpha}), which back-substitutes into (\ref{eq.linep}) for closed-form translation. This two-point minimal solution enables efficient deterministic search.}

\textcolor{black}{\textbf{Decoupled Two-Stage Search.} The minimal solution enables deterministic convergence through staged optimization:}

\textcolor{black}{\textbf{Stage 1: Yaw estimation.} Using TIMs from all correspondence pairs, we formulate consensus maximization over 1DoF yaw space:}
\begin{equation}\label{eq.costr}
\max_{\alpha,\{z_{ij}\}}\sum z_{ij} \quad \text{s.t.} \quad z_{ij} \|d_{ij}(\alpha)\| \leq n_{ij}
\end{equation}
\textcolor{black}{where $z_{ij}\in\{0,1\}$ indicates inlier status for pair $(i,j)$, and $n_{ij}$ bounds TIM noise. We perform exhaustive discretized search over $\alpha \in [-\pi, \pi]$ using inlier voting to find globally optimal $\hat{\alpha}$, eliminating most outliers.}

\textcolor{black}{\textbf{Stage 2: Translation estimation.} With rotation fixed, we estimate translation from filtered inliers via consensus maximization:}
\begin{equation}\label{eq.costt}
\max_{\mathbf{t},\{z_i\}}\sum z_i \quad \text{s.t.} \quad z_i \|\mathbf{u}_i - \pi(\mathbf{R}(\hat{\alpha})\mathbf{F}_i+\mathbf{t},\mathcal{K})\| \leq n_i
\end{equation}
\textcolor{black}{where $\pi(\cdot)$ is camera projection. We formulate this 3DoF problem as maximum clique detection in a geometric compatibility graph, yielding globally optimal $\hat{\mathbf{t}}$ via branch-and-bound search.}


\textcolor{black}{The complete algorithm proceeds as follows. We first extract 2D-3D correspondences through place recognition (NetVLAD ~\cite{NetVLAD}) and feature matching (SuperPoint + LightGlue ~\cite{detone2018superpoint, lightglue}), then exhaustively search 1DoF yaw via (\ref{eq.costr}). Subsequently, we search 3DoF translation via (\ref{eq.costt}) on filtered inliers, and finally refine through nonlinear optimization. Decoupling 6DoF into 1DoF+3DoF ensures polynomial-time global optimality: $O(K^2)$ for rotation voting over $K$ candidate estimates, plus $O(N^3)$ for translation clique search over $N$ remaining inliers after filtering. In practice, this completes in 0.1-0.7 seconds for 15-150 correspondences, as demonstrated in Table 6. The efficiency stems from exploiting vehicle motion constraints to reduce search dimensionality while preserving optimality guarantees.}

\vspace{-0.2cm}
\subsubsection{Online matching module}\label{sec:online_match}
\textcolor{black}{While the initialization module achieves globally optimal pose matching in seconds, continuous localization requires different trade-offs. To obtain frequent map observations for drift correction, the system must prioritize computational efficiency over global optimality. We therefore adopt a RANSAC-based approach that balances robustness and real-time performance.}


\textcolor{black}{\textbf{Challenge: Extreme outliers in long-term operation.} Feature matching across temporally distant sessions faces severe challenges from environmental variations. Our experiments with the ZJG and NC datasets reveal that vegetation growth, ongoing construction, seasonal changes, and varying viewpoints can produce outlier rates exceeding 80\%, as illustrated in Fig.\ref{fig.matchcases}(a)-(c). Under such extreme conditions, traditional RANSAC-based methods struggle to converge within real-time constraints.}


\textcolor{black}{\textbf{IMU-aided 2-point minimal solution.} We reduce the minimal required correspondences from 3-4 points to just 2 points by leveraging inertial measurements. As derived in Sec.~\ref{sec:init}, gravity alignment from the IMU enables a 4DoF pose formulation, and the collinearity constraint~\eqref{eq.linep} yields a closed-form 2-point minimal solution. The success probability $P$ of RANSAC after $k$ iterations fundamentally depends on the minimal sample size $n$ and inlier rate $w$:}
\begin{equation}
1-P = (1-w^{n})^{k}
\label{eq:success_rate}
\end{equation}
\textcolor{black}{This relationship reveals that reducing $n$ from 3 to 2 dramatically improves convergence under high outlier rates. With fixed iterations, our 2-point formulation achieves substantially higher success probability than traditional 3-4 point methods when $w \leq 0.2$, enabling robust operation even when outliers dominate.}

\textcolor{black}{To further improve the effective inlier rate $w$ in Eq.~\eqref{eq:success_rate}, we implement a feature scoring network that predicts correspondence reliability based on descriptor patterns. This learned metric guides RANSAC sampling toward higher-quality matches, increasing the probability of selecting inlier pairs in each iteration. Experimental results (Sec.~\ref{sec:online_match_eval}) demonstrate that the proposed 2P RANSAC framework maintains robust performance under outlier rates from 60\% to 80\%, conditions where traditional methods fail.}

\textcolor{black}{\textbf{Multi-camera extension (MC2P).} We extend the 2-point solution to leverage observations from multiple cameras through our MC2P algorithm. The approach consolidates multi-view observations by independently extracting and matching features in each camera view against retrieved map keyframes, transforming all 2D-3D correspondences into a unified reference frame using calibrated extrinsics, and applying the 2-point RANSAC on the combined correspondence set. This multi-camera fusion provides dual benefits: increasing the absolute number of inliers available for sampling (improving $w$ in Eq.~\eqref{eq:success_rate}), and capturing complementary scene geometry where side cameras observe structural features that forward-facing cameras miss, enabling successful pose estimation even when individual views provide insufficient constraints.}


\textcolor{black}{The online matching module operates at camera frame rate: for each query frame, place recognition retrieves candidate map keyframes, feature extraction and scoring identify correspondences, and pose estimation proceeds via 2P or MC2P RANSAC with 100 iterations. The reduced minimal sample size ($n=2$ versus 3-4) enables significantly faster convergence under high outlier rates compared to traditional methods. This computational efficiency, achieved by trading the initialization module's global optimality for probabilistic convergence, enables continuous drift correction at high frequency throughout multi-kilometer trajectories.}

\vspace{-0.2cm}
\subsubsection{Map-based VINS module}\label{sec:vins}
\textcolor{black}{The map-based VINS module fuses local VIO estimates with map observations from the initialization and online matching modules, providing real-time, causal state estimation with bounded error. This section details the filter-based framework that integrates multiple isolated maps through state augmentation.}



\vspace{0.2cm}
\textcolor{black}{\textbf{State Augmentation for Multi-Map Fusion.} We extend the Open-VINS state vector~\cite{openvins} to simultaneously track transformations to $N$ isolated maps:}
\begin{align}
    \mathbf{x}_{t}&=\begin{bmatrix}\mathbf{x}_{VIO_t}^{\top}&\mathbf{x}_{SW_t}^{\top}&\mathbf{x}_{\tau_t}^{\top}&\mathbf{x}_{KF_t}^{\top}\end{bmatrix}^{\top}\label{eq:state}\\
    \mathbf{x}_{VIO_t}&=\begin{bmatrix}{}^{I_t}\mathbf{q}_{L}^{\top}&{}^{L}\mathbf{v}_{I_t}^{\top}&{}^{L}\mathbf{p}_{I_t}^{\top}&\mathbf{b}_{g_t}^{\top}&\mathbf{b}_{a_t}^{\top}\end{bmatrix}^{\top}\notag\\
    \mathbf{x}_{SW_t}&=\begin{bmatrix}{}^{I_{t-1}}\mathbf{q}_{L}^{\top}&{}^{L}\mathbf{p}_{I_{t-1}}^{\top}&\cdots&{}^{I_{t-B}}\mathbf{q}_{L}^{\top}&{}^{L}\mathbf{p}_{I_{t-B}}^{\top}\end{bmatrix}^{\top}\notag\\
    \mathbf{x}_{\tau_t}&=\begin{bmatrix}{}^{G^{1}_t}\mathbf{q}_{L}^{\top}&{}^{G^{1}_t}\mathbf{p}_{L}^{\top}&\cdots&{}^{G^{N}_t}\mathbf{q}_{L}^{\top}&{}^{G^{N}_t}\mathbf{p}_{L}^{\top}\end{bmatrix}^{\top} \notag\\
    \mathbf{x}_{KF_t}&=\begin{bmatrix}\cdots{}^{G^{1}_t}\mathbf{q}_{kf_j}^{\top}\!&\!{}^{G^{1}_t}\mathbf{p}_{kf_j}^{\top}\cdots{}^{G^{N}_t}\mathbf{q}_{kf_k}^{\top}\!&\!{}^{G^{N}_t}\mathbf{p}_{kf_k}^{\top}\cdots\end{bmatrix}\!^{\top} \notag
\end{align}


\textcolor{black}{where $\mathbf{x}_{VIO_t}$ contains the standard VIO state: IMU pose ${}^{L}\mathbf{T}_{I_t}\triangleq\{{}^{I_t}\mathbf{q}_{L},{}^{L}\mathbf{p}_{I_t}\}$, velocity ${}^{L}\mathbf{v}_{I_t}$, and IMU biases $\mathbf{b}_{g_t}, \mathbf{b}_{a_t}$. The sliding window $\mathbf{x}_{SW_t}$ maintains $B$ cloned historical poses for multi-view feature triangulation, with efficient marginalization as the window advances.}
Suppose we have $N$ isolated pre-built maps, then the elements ${}^{G^{i}_t}\mathbf{q}_{L},\, {}^{G^{i}_t}\mathbf{p}_{L}, \,i=1\cdots N$ represent the relative transformation between the local VIO frame $\{L\}$ and the $i$-th map frame $\{G^{i}\}$, i.e., the so-called augmented variable. $\mathbf{x}_{KF_t}$ contains a set of keyframe poses from $N$ isolated maps. For instance, ${}^{G^{1}_t}\mathbf{q}_{kf_j}$ and ${}^{G^{1}_t}\mathbf{p}_{kf_j}$ represent the orientation and the position of the $j$-th keyframe from the $1$st map, respectively.


\vspace{0.2cm}
\textcolor{black}{\textbf{State Propagation.} The augmented system evolves according to IMU-driven kinematics:}
\begin{equation}\label{eq:kinematics}
\left\{\begin{aligned} ^{I_t}\dot{\mathbf{q}}_{L}&=\frac{1}{2}\Omega(\boldsymbol{\omega}_{m_t}-\mathbf{b}_{g_t}-\mathbf{n}_{g_t}){}^{I_t}\mathbf{q}_{L}&\\
    ^{L}\dot{\mathbf{v}}_{I_t}&=C(^{I_t}{\mathbf{q}}_{L})^{\top}(\mathbf{a}_{m_t}-\mathbf{b}_{a_t}-\mathbf{n}_{a_t})+\mathbf{g}\\
    ^{L}\dot{\mathbf{p}}_{I_t}&={}^{L}\mathbf{v}_{I_t}\\
    \dot{\mathbf{b}}_{g_t}&=\mathbf{n}_{wg_t}\;\;\;\,
    \dot{\mathbf{b}}_{a_t}=\mathbf{n}_{wa_t}\\
    ^{G_t^{i}}\dot{\mathbf{q}}_{L}&=\mathbf{0}\;\;\;\;\;\;
    ^{G_t^{i}}\dot{\mathbf{p}}_{L}=\mathbf{0}\\
    ^{G_t^{i}}\dot{\mathbf{q}}_{kf_j}&=\mathbf{0}\;\;\;\;
    ^{G_t^{i}}\dot{\mathbf{p}}_{kf_j}=\mathbf{0}\\
    \end{aligned}
    \right. 
\end{equation}
\textcolor{black}{where standard VIO components evolve through IMU integration while map-related variables remain static. Here $C(\cdot)$ converts quaternion to rotation matrix, $\boldsymbol{\omega}_{m_t}$ and $\mathbf{a}_{m_t}$ are gyroscope and accelerometer measurements, $\mathbf{g}=\begin{bmatrix}0&0&-9.81\end{bmatrix}^{\top}$ m/s$^2$ is gravity, and noise terms follow standard conventions~\cite{msckf}. The key property preserving causality: map-to-local transformations ${}^{G^{i}_t}\mathbf{T}_{L}$ remain constant, and we correct VIO drift by adjusting these transformations rather than retroactively modifying historical poses. At each timestep, the filter outputs ${}^{G^{i}}\mathbf{T}_{I_t} = {}^{G^{i}_t}\mathbf{T}_{L} \cdot {}^{L}\mathbf{T}_{I_t}$ using only past observations.}

\textcolor{black}{Discretizing (\ref{eq:kinematics}) yields standard EKF propagation $\mathbf{x}_{t+1|t}=\mathbf{f}(\mathbf{x}_{t},\mathbf{a}_{m_t}-\mathbf{n}_{a_t},\boldsymbol{\omega}_{m_t}-\mathbf{n}_{\omega_t})$ with covariance update $\mathbf{P}_{t+1|t}=\mathbf{\Phi}_{t+1|t}\mathbf{P}_{t|t}\mathbf{\Phi}_{t+1|t}^{\top}+\mathbf{G}_{t+1|t}\mathbf{Q}_{t+1}\mathbf{G}_{t+1|t}^{\top}$, where Jacobians $\mathbf{\Phi}_{t+1|t}$ and $\mathbf{G}_{t+1|t}$ follow standard MSCKF derivations~\cite{msckf,openvins}.}

\vspace{0.2cm}
\textcolor{black}{\textbf{Observation Integration via Null-Space Projection.}} As indicated in Fig.\ref{fig:frame}, there are two kinds of observation functions, the local feature based (the \textcolor{black}{blue} shade part) and the map feature based (the \textcolor{black}{pink} shade part). \textcolor{black}{Local features $^{L}\mathbf{F}_{i}$ tracked across sliding window frames provide VIO constraints:}
\begin{equation}\label{eq:local ob}
    {}^{loc}\mathbf{y}_t^{i} = \mathrm{h}({}^{I_t}\mathbf{R}_{L}({}^{L}\mathbf{F}_{i_t}-{}^{L}\mathbf{p}_{I_t}))+{}^{loc}\boldsymbol{\gamma}_{t}^{i}
\end{equation}
\textcolor{black}{where $\mathrm{h}(\cdot)$ projects 3D points to the image plane and ${}^{loc}\boldsymbol{\gamma}_{t}^{i}$ is measurement noise\footnote{\textcolor{black}{For non-identity camera-IMU extrinsic ${}^{C}\mathbf{T}_{I}$, this becomes ${}^{loc}\mathbf{y}_t^{i} = \mathrm{h}({}^{C}\mathbf{R}_{I}{}^{I_t}\mathbf{R}_{L}({}^{L}\mathbf{F}_{i_t}-{}^{L}\mathbf{p}_{I_t})+{}^{C}\mathbf{p}_{I})+{}^{loc}\boldsymbol{\gamma}_{t}^{i}$, where the constant rotation ${}^{C}\mathbf{R}_{I}$ does not affect observability analysis.}}. These observations constrain VIO state but accumulate drift.}

\textcolor{black}{Map features $^{G^k}\mathbf{F}_{i}$ from the $k$-th map provide drift correction through observations linking: (i) the current camera, (ii) keyframes in the same map $^{G^k}KF_j$, and (iii) keyframes in other maps $^{G^s}KF_l$ when features span multiple maps:}
\begin{equation}\label{eq:map ob in cam}
^{cam}\mathbf{y}_{t}^{ki}\!=\!\mathrm{h}\!\left[{}^{I_t}\mathbf{R}_{L}({}^{G_t^k}\mathbf{R}_{L}^{\top}({}^{G^k}\mathbf{F}_{i_t}\!\!-\!\!{}^{G_t^k}\mathbf{p}_{L})-{}^{L}\mathbf{p}_{I_t})\right]+{}^{cam}\boldsymbol{\gamma}_{t}^{ki}
\end{equation}
\begin{equation}\label{eq:map ob in kf1}
^{{}^{G^k}KF_{j}}\mathbf{y}_{t}^{ki}=\mathrm{h}\left[{}^{G_t^{k}}\mathbf{R}_{KF_j}^{\top}({}^{G^k}\mathbf{F}_{i_t}-{}^{G^k_t}\mathbf{p}_{KF_j})\right]+{}^{{}^{G^k}KF_{j}}\boldsymbol{\gamma}_{t}^{ki}
\end{equation}
\begin{equation}\label{eq:map ob in kf2}
\begin{aligned}
^{{}^{G^s}KF_{l}}\mathbf{y}_{t}^{ki}&=\mathrm{h}\Big[{}^{G_t^{s}}\mathbf{R}_{KF_l}^{\top}\big[{}^{G_t^s}\mathbf{R}_{L}{}^{G_t^k}\mathbf{R}_{L}^{\top}({}^{G^k}\mathbf{F}_{i_t}-{}^{G^k_t}\mathbf{p}_{L})\\
&\;\;\;\;+{}^{G_t^s}\mathbf{p}_{L}-{}^{G_t^s}\mathbf{p}_{KF_l}\big]\Big]+{}^{{}^{G^s}KF_{l}}\boldsymbol{\gamma}_{t}^{ki}
\end{aligned}
\end{equation}

\textcolor{black}{Equation (\ref{eq:map ob in cam}) transforms map features to the current camera via estimated transformation ${}^{G^k_t}\mathbf{T}_{L}$. Equation (\ref{eq:map ob in kf1}) provides intra-map constraints. Equation (\ref{eq:map ob in kf2}) enables cross-map constraints when features appear in multiple maps, critical for multi-map fusion.}

\textcolor{black}{Following MSCKF~\cite{msckf}, we exclude 3D feature positions from the state vector to maintain tractable dimensionality. Instead, features are marginalized through null-space projection. For local features observed across $n\geq 2$ frames, the stacked Jacobian $\mathbf{H}^{i}_{{}^{L}\mathbf{F}_{i}}$ has dimensions $2n\times 3$. The left null-space matrix $\mathbf{N}^{\top}_{{}^{L}\mathbf{F}_{i}}$ satisfies $\mathbf{N}^{\top}_{{}^{L}\mathbf{F}_{i}}\mathbf{H}^{i}_{{}^{L}\mathbf{F}_{i}}=\mathbf{0}$, eliminating feature dependency from the update. Map features similarly require at least two observations (current camera + map keyframe) to ensure the row dimension of the stacked Jacobian exceeds its column dimension, guaranteeing null-space existence.}

\vspace{0.2cm}
\textcolor{black}{\textbf{Filter Update via Schmidt-EKF.} We partition the state into active $\mathbf{x}_{A_t}=[\mathbf{x}_{VIO_t}^{\top}\;\mathbf{x}_{SW_t}^{\top}\;\mathbf{x}_{\tau_t}^{\top}]^{\top}$ and nuisance $\mathbf{x}_{N_t}=\mathbf{x}_{KF_t}$ components. For local features, standard MSCKF update applies to $\mathbf{x}_{A_t}$. For map features, after null-space projection yields residual $\mathbf{r}^{*}_t = \mathbf{H}^{*}_{A_t}\tilde{\mathbf{x}}_{A_t} + \mathbf{H}^{*}_{N_t}\tilde{\mathbf{x}}_{N_t} + \boldsymbol{\gamma}^{*}_t$, we employ Schmidt-EKF update:}
\textcolor{black}{\begin{align}
\hat{\mathbf{x}}_{A_t} &= \hat{\mathbf{x}}_{A_t|t-1} + \mathbf{K}_{A_t}\mathbf{r}^{*}_t,\quad 
\hat{\mathbf{x}}_{N_t} = \hat{\mathbf{x}}_{N_t|t-1} \label{eq:schmidt_update}\\
\mathbf{K}_{A_t} &= (\mathbf{P}_{AA_{t|t-1}}\mathbf{H}^{*\top}_{A_t} + \mathbf{P}_{AN_{t|t-1}}\mathbf{H}^{*\top}_{N_t})\mathbf{S}^{-1}_t \notag
\end{align}}
\textcolor{black}{where $\mathbf{S}_t = \mathbf{H}^{*}_{t}\mathbf{P}_{t|t-1}\mathbf{H}^{*\top}_{t} + \mathbf{R}_t$. Crucially, map keyframes $\mathbf{x}_{N_t}$ remain unchanged, reducing computational complexity from $O(n_k^2)$ to $O(n_k)$ while maintaining conservative yet consistent estimates~\cite{schmidtEKF}. Our high-precision offline mapping (Sec.~\ref{sec:offline_map}) ensures that map reconstruction errors are negligible compared to VIO drift and matching uncertainties, justifying this fixed-map treatment without degrading localization performance.}

\textcolor{black}{The complete filter cycle operates as follows: (a) IMU measurements propagate VIO state via (\ref{eq:kinematics}), (b) visual observations from both local features (\ref{eq:local ob}) and map features (\ref{eq:map ob in cam})-(\ref{eq:map ob in kf2}) are processed with null-space projection to eliminate 3D feature dependencies, (c) filter update corrects VIO state and map-to-local transformations using standard MSCKF for local features and Schmidt-EKF for map features. This architecture achieves three critical properties: causality is preserved as no future information is required, bounded localization error is maintained through continuous map constraints that prevent drift accumulation, and multi-map scalability} is realized via augmented state $\mathbf{x}_{\tau_t}$ tracking multiple isolated maps.

\section{SYSTEM EVALUATION METRIC}\label{sec:metrics}

\begin{figure}[tp]
  \centering
 \setlength{\belowcaptionskip}{-0.08cm}
 \includegraphics[width=0.48\textwidth]{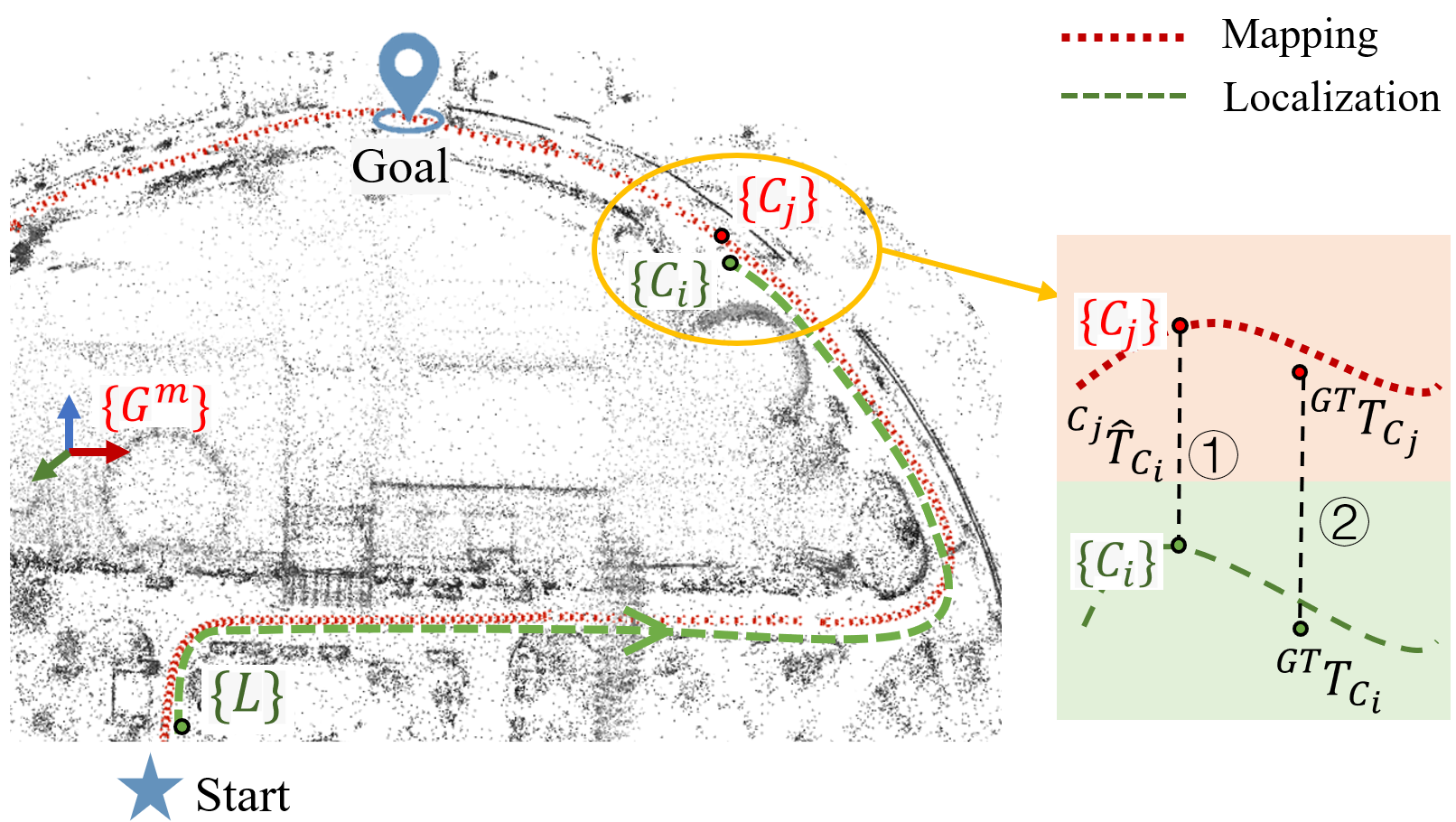}
  \caption{\textcolor{black}{Error decomposition for map-based robot navigation.
  \textbf{Task}: Navigate from start to goal in map frame $\{G^m\}$.
  \textbf{Process}: The system first aligns local frame $\{L\}$ with map frame through initialization, then matches current image $C_i$ with map image $C_j$ at each timestep $i$ to update pose $^{G^m}\hat{\mathbf{T}}_{C_i}$ for control.}
  During the robot's navigation process, take timestamp $i$ as an example, the error analysis between $C_i$ and the retrieved map frame $C_j$ is highlighted in the bottom right corner.}
  \label{fig:evaluation_system}
\vspace{-0.4cm}
\end{figure}

\textcolor{black}{\textbf{The Control-Evaluation Mismatch.} Robot control loops require real-time pose estimates at each decision instant. However, standard SLAM evaluation metrics (ATE, RPE) measure accuracy of post-processed trajectories after global optimization and loop closure corrections. This creates a fundamental mismatch: the poses being evaluated are not the poses that controlled the robot. While these metrics effectively assess mapping and SLAM performance, they inadequately reflect the causal estimation quality required for robot control applications.}

\textcolor{black}{Consider the navigation scenario in Fig.\ref{fig:evaluation_system}: A robot navigates from start to goal using map-based localization. At time $i$, the system matches current image $C_i$ with retrieved map frame $C_j$ to estimate pose $^{G^m}\hat{\mathbf{T}}_{C_i}$ for control feedback. Traditional metrics evaluate this pose after potential future corrections (e.g., loop closures at time $j > i$), but the controller operated using the uncorrected estimate available at time $i$. This delayed correction arrives too late for decisions already executed. Additionally, metrics like ATE perform global trajectory alignment using all poses, which incorporates future information unavailable during real-time operation. While RPE avoids global alignment through local relative measurements, it typically evaluates retrospectively optimized trajectories rather than the real-time estimates seen by controllers. For practical deployment, initialization success and matching accuracy under long-term appearance variations are equally critical yet invisible to trajectory-only metrics.}

\textcolor{black}{To properly evaluate localization systems for robot control applications, we need metrics that reflect the causal nature of control loops and separately assess each stage of the localization pipeline. Traditional trajectory metrics provide overall accuracy but cannot isolate whether errors originate from map quality, feature matching failures, or tracking drift. We therefore decompose the total localization error into four components corresponding to distinct sources, each evaluated independently to diagnose system behavior:}
\textcolor{black}{\begin{itemize}
\item \textbf{Mapping error} ($E^k_{mapping}$, $E^p_{mapping}$): Quantifies the geometric accuracy of the pre-built map itself, including both keyframe poses and triangulated feature points. This establishes the accuracy ceiling for all subsequent localization.
\item \textbf{Matching error} ($E^t_{alignment}$, $E^r_{alignment}$): Measures the precision of associating current observations with map frames, covering both initial map alignment at startup and continuous frame-to-map matching during operation. High matching error directly degrades observation quality in pose estimation.
\item \textbf{Local-based trajectory error} ($E^{local}_{trajectory}$): Evaluates causal tracking performance by measuring the accuracy of real-time pose estimates in the local frame. These estimates use only past information without retrospective corrections, directly reflecting the feedback quality available to control loops.
\item \textbf{Map-based trajectory error} ($E^{map}_{trajectory}$): Assesses end-to-end localization accuracy in the map frame when initialization succeeds. This combines all error sources and is applicable in scenarios where all methods successfully initialize.
\end{itemize}}

\textcolor{black}{Fig.\ref{fig:evaluation_system} illustrates how these components manifest in robot navigation. The following subsections detail the measurement methodology for each component.}

\subsection{MAPPING ACCURACY EVALUATION}

\textcolor{black}{Map quality directly affects downstream localization performance, as inaccurate keyframe poses or feature positions propagate errors to matching and tracking stages. We assess mapping accuracy through two complementary metrics.}

\vspace{-0.2cm}
\subsubsection{Keyframe Pose Accuracy}
\textcolor{black}{We evaluate the geometric accuracy of map keyframe poses using the Absolute Trajectory Error (ATE) metric. After aligning the map trajectory with ground truth via least squares in $SE(3)$, we compute the root mean square error of position deviations:}
\begin{equation}\label{eq:map_kf}
    E_{mapping}^{k} = \sqrt{\frac{1}{N_1}\sum_{i=1}^{N_1} \|\mathbf{E}_i\|^{2}} 
\end{equation}
\textcolor{black}{where the position error at keyframe $i$ is:}
\begin{equation}
    \mathbf{E}_i = {^{GT}\mathbf{F}_i} - {^{GT}\mathbf{T}_{G^m}} {^{G^m}\mathbf{F}_i}
\end{equation}
\textcolor{black}{Here $^{GT}\mathbf{F}_i$ and $^{G^m}\mathbf{F}_i$ denote ground truth and map keyframe positions at time $i$ ($i{=}1,{\ldots},N_1$), and ${^{GT}\mathbf{T}_{G^m}}$ is the rigid-body transformation obtained via least squares alignment. For notational convenience, we define:}
\begin{equation}
    {^{A}\mathbf{T}_{B}}\mathbf{F} \triangleq {^{A}\mathbf{R}_{B}} \mathbf{F} + {^{A}\mathbf{t}_{B}}
\end{equation}

\vspace{-0.2cm}
\subsubsection{Map Point Reconstruction Accuracy}
\textcolor{black}{Since direct point-to-point correspondence between visual reconstruction and ground truth is unavailable, we apply ICP to align the reconstructed point cloud with ground truth, then compute nearest-neighbor matching error. Let $\mathbf{M}$ and $\mathbf{Q}$ denote ground truth and reconstructed point sets, with $\mathbf{m}_j$ and $\mathbf{q}_j$ being the $j$-th corresponding pair ($j{=}1,{\ldots},N_2$):}
\begin{equation}\label{eq:map_pt}
    E_{mapping}^p = \sqrt{\frac{1}{N_2}\sum_{j=1}^{N_2} \|\mathbf{m}_j - \mathbf{q}_j\|^{2}}
\end{equation}
\textcolor{black}{Note that this metric provides an overall assessment of map reconstruction quality. Due to the highly non-Gaussian error distribution of visual point clouds and inherent ICP alignment limitations with heterogeneous point densities, we use this metric to verify that reconstruction is not severely degraded, rather than claiming precise point-wise accuracy.}


\subsection{MATCHING ACCURACY EVALUATION}\label{sec:matching_eval}
\textcolor{black}{Matching accuracy quantifies the precision of associating current observations with map frames. This evaluation applies to both initial map alignment at system startup and online frame-to-map matching during navigation.}

\textcolor{black}{Consider matching between current camera frame $\{C_i\}$ at time $i$ and a retrieved map keyframe $\{C_j\}$ (as illustrated in Fig.\ref{fig:evaluation_system}, bottom-right). Ground truth trajectories from external sensors provide reference poses $^{GT}\mathbf{T}_{C_i}$ and $^{GT}\mathbf{T}_{C_j}$. The ground truth relative transformation is:}
\begin{equation}
^{{C_{j}}}\mathbf{T}_{{C}_{i}} = ({^{GT}}\mathbf{T}_{C_{j}})^{-1} \; {^{GT}}\mathbf{T}_{C_{i}}
\end{equation}
\textcolor{black}{We separately measure translation and rotation errors between the estimated transformation $^{C_{j}}\hat{\mathbf{T}}_{C_{i}}$ and ground truth $^{C_{j}}\mathbf{T}_{C_{i}}$:}
\begin{align}
    E_{alignment}^{t} &= \|^{C_{j}}\mathbf{t}_{C_{i}} - {^{C_{j}}\hat{\mathbf{t}}_{C_{i}}}\| \label{eq:alignt} \\
    E_{alignment}^{r} &= \arccos\left(\frac{\text{tr}(^{C_{j}}\hat{\mathbf{R}}_{C_{i}}^{\top} {^{C_{j}}\mathbf{R}_{C_{i}}}) - 1}{2}\right) \label{eq:alignr}
\end{align}
\textcolor{black}{For initialization, this assesses the accuracy of the globally optimal alignment $^{G^m}\mathbf{T}_L$ established at startup. For online matching, it evaluates per-frame association accuracy, which directly affects map observation quality in the subsequent filter update.}

\subsection{LOCAL-BASED TRAJECTORY ACCURACY}\label{sec:local_traj_eval}
\textcolor{black}{Robot control loops operate on real-time pose estimates available at each decision instant. However, standard SLAM evaluation metrics measure accuracy of trajectories after global optimization, including loop closure corrections. This creates a mismatch: the poses being evaluated are not the poses that actually controlled the robot.}

\textcolor{black}{Consider a concrete scenario: at time $t{=}10$s, the controller receives pose estimate $\mathbf{p}_{10}$ and uses it for navigation decisions. At $t{=}20$s, loop closure detection triggers optimization that revises the $t{=}10$s pose to $\mathbf{p}'_{10}$. The ATE metric evaluates error using $\mathbf{p}'_{10}$ after computing optimal alignment over the entire trajectory. While RPE measures relative pose changes without global alignment, it still evaluates the loop-closed trajectory where earlier poses have been retrospectively corrected. Both metrics therefore assess post-processed results rather than the causal estimates that guided control.}

\textcolor{black}{To measure control-relevant accuracy, we evaluate the real-time trajectory before any retrospective corrections. Starting with the estimated trajectory $\{^{L}\mathbf{x}_i | ^{L}\mathbf{x}_i\in SE_3, i=1,\cdots,N_3\}$ and ground truth $\{^{GT}\mathbf{x}_i | ^{GT}\mathbf{x}_i\in SE_3, i=1,\cdots,N_3\}$, we align only the first frame to obtain ${^{GT}\mathbf{T}_{L}}$ by aligning $^{L}\mathbf{x}_1$ with $^{GT}\mathbf{x}_1$, then measure all subsequent errors using this fixed transformation:}
\begin{equation}\label{eq.local_traj}
    {E}^{local}_{trajectory} = \sqrt{\frac{1}{N_3-1}\sum_{i=2}^{N_3} \Vert ^{GT}\mathbf{F}_i - {^{GT}\mathbf{T}_{L}} ^{L}\mathbf{F}_i \Vert^{2}}
\end{equation}
\textcolor{black}{We align only at the first frame because initialization accuracy is measured separately in Sec.~\ref{sec:matching_eval}. This ensures that the error at time $i$ reflects only the information available up to time $i$, matching the causality constraint of robot control.}

\subsection{MAP-BASED TRAJECTORY ACCURACY}\label{sec:map_traj_eval}
\textcolor{black}{In scenarios where all evaluated methods successfully complete initialization, we can directly compare localization accuracy in the map frame. Since each method automatically performs map alignment during its initialization phase, we evaluate the estimated trajectory $\{^{G^m}\mathbf{F}_i | i{=}1,{\ldots},N_3\}$ directly against ground truth:}
\begin{equation}\label{eq.map_traj}
    {E}^{map}_{trajectory} = \sqrt{\frac{1}{N_3}\sum_{i=1}^{N_3} \Vert ^{GT}\mathbf{F}_i - {^{G_m}\mathbf{F}_i} \Vert^{2}}
\end{equation}
\textcolor{black}{This is equivalent to ATE without post-hoc alignment. The metric reflects the combined effect of mapping, matching, and tracking errors, making it useful for overall system comparison in favorable scenarios.}

\section{EXPERIMENTS}\label{sec:Experiments}

\textcolor{black}{This section provides comprehensive experimental validation of the proposed multi-camera multi-map VILO system through systematic evaluation of its core components and overall performance. The experimental design follows a modular approach that progresses from individual component verification to integrated system assessment. This hierarchical validation strategy ensures that both the underlying algorithmic components and the complete system meet the requirements for practical robot control applications, where real-time, causal state estimation is critical.}

\textcolor{black}{The experimental validation is organized into five complementary aspects: (a) Mapping accuracy and robustness (Sec.~\ref{sec:mapping_eval}) evaluates the precision of constructed maps and validates the impact of multi-sensor integration on reconstruction quality; (b) Matching accuracy and robustness (Sec.~\ref{sec:matching_eval}) assesses the reliability of our minimal solvers and deterministic initialization under challenging environmental conditions; (c) Localization accuracy and robustness (Sec.~\ref{sec:localization_eval}) validates multi-map fusion capabilities and multi-camera configurations in diverse scenarios; and (d) Real-time system evaluation (Sec.~\ref{sec:realtime}) compares integrated system performance against established visual-inertial SLAM methods in terms of causal trajectory accuracy suitable for robot control.}

\subsection{Datasets}  

\textcolor{black}{Our experimental validation employs three complementary datasets that span diverse operational scenarios and environmental conditions.}

\textbf{Newer College Dataset (NC)~\cite{oxford-multicam-dataset}:} A handheld multi-camera LiDAR inertial dataset covering diverse indoor and outdoor scenes (\textit{math} and \textit{quad}) over a 4.5 km trajectory. Its multi-camera setup and varied environments are suitable for testing mapping, matching, and localization modules.

\textcolor{black}{\textbf{EuRoC MAV Dataset (EuRoC)~\cite{EuRoC}:} A stereo camera and IMU dataset collected from a micro aerial vehicle operating in two indoor environments: a machine hall with industrial infrastructure and smaller Vicon rooms with controlled motion capture ground truth. This widely-used benchmark facilitates direct performance comparison with existing visual-inertial methods.}

\begin{figure}[t] %
	\setlength{\belowcaptionskip}{-0.4cm}
	\begin{center}
		\includegraphics[width=0.45\textwidth]{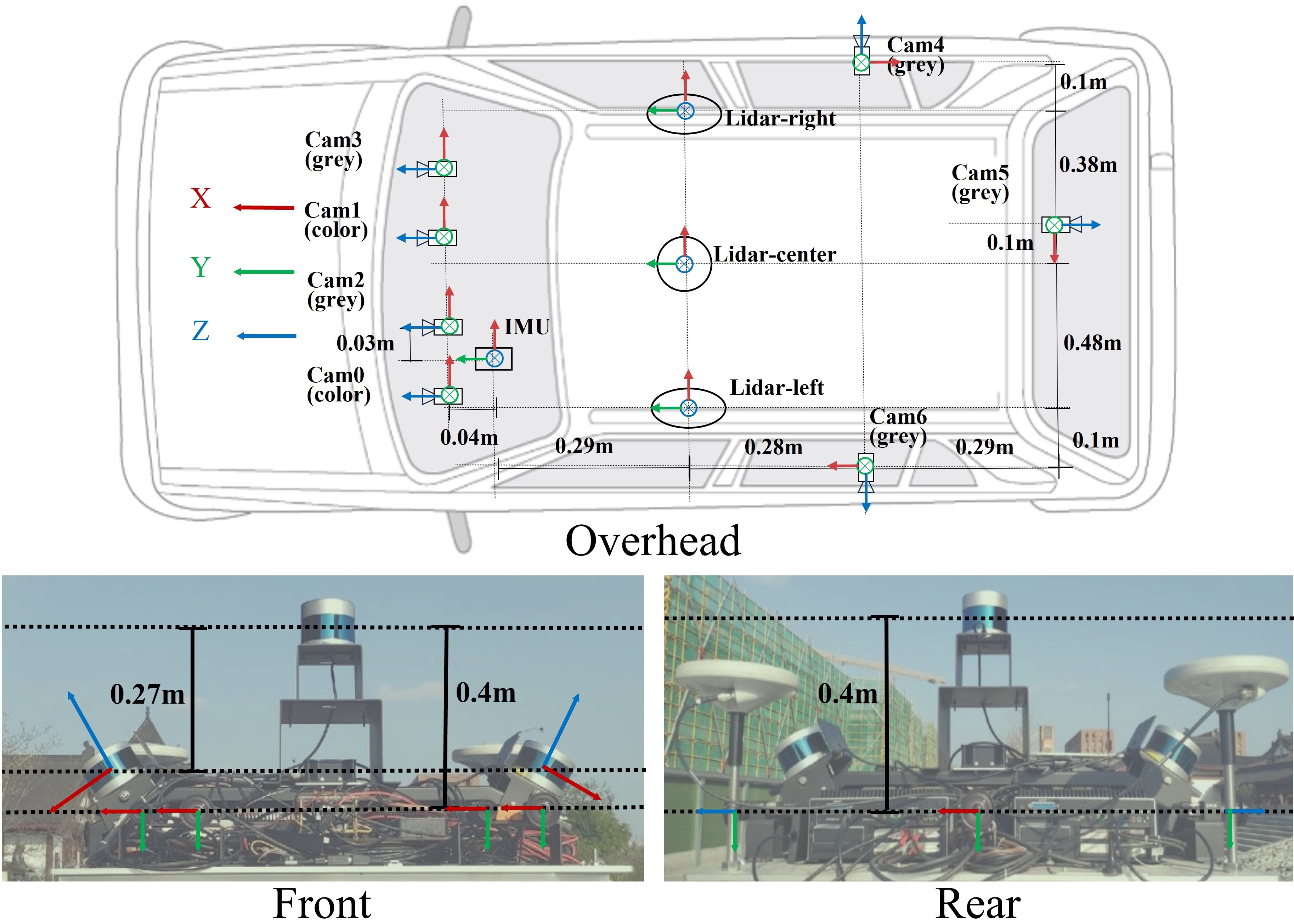}
		\caption{Multi-sensor data collection vehicle used for the ZJG dataset.}
		\label{fig:car_setup}
	\end{center}
\vspace{-0.4cm}
\end{figure}

\textcolor{black}{\textbf{ZJG Campus Dataset (ZJG):} To rigorously evaluate the system's performance under the target conditions of long-term operation in large-scale, dynamic, and unstructured environments, we collected a dedicated multi-session dataset on the Zijingang Campus of Zhejiang University. Data acquisition was conducted using a vehicle-mounted multi-sensor platform (Fig.\ref{fig:car_setup}) over a nine-month period, capturing 265,000 m² of operational area with trajectories exceeding 55 km. The vehicle platform ensures consistent sensor geometry and temporal synchronization across multiple cameras providing surround-view coverage, a high-grade IMU, and LiDAR sensors for ground truth generation (Table~\ref{tab:sensor_config}). This dataset intentionally encompasses significant environmental variations that challenge visual localization: seasonal appearance changes driven by vegetation growth and snow coverage, varying illumination from full daylight to dusk conditions, structural modifications from ongoing campus construction, dynamic elements including pedestrians and vehicles, and substantial viewpoint variations between collection sessions. Hardware-synchronized sensor streams enable precise temporal alignment, while high-accuracy ground truth trajectories were generated using FAST-LIO2~\cite{fastlio2} processed offline on LiDAR-inertial data. These characteristics render the ZJG dataset particularly demanding for feature matching and long-term localization, providing a rigorous testbed for evaluating robustness claims in unstructured, dynamic environments. The dataset, along with the system, is released as an open-source resource\url{https://anonymous.4open.science/r/Multi-cam-Multi-map-VILO-7993}. Further details regarding this dataset are provided in the Supplementary Material.}

\begin{table}[t]
    \centering
    \footnotesize
    \caption{Sensor configurations on the data collection vehicle.}
    \label{tab:sensor_config}
    \resizebox{0.5\textwidth}{!}{
        \begin{tabular}{cccc}
            \hline\hline
            Sensor & Type & Hz & Specification\\
            \hline
            Front Color Camera & BFS-PGE-31S4C-C & 16 & $2048\times 1536$ $^1$ \\
            Front Grey Camera & BFLY-PGE-03S3M-CS & 16 & $648\times 488$ $^1$ \\
            Left\&Right Fisheye & BFLY-PGE-03S3M-CS & 16 & $648\times 488$ $^1$ \\
            3D LiDAR(Top) & VLP-32C & 10 & $32$ $^2$ \\
            3D LiDAR(Side) & VLP-16 & 10 & $16$ $^2$ \\
            IMU & Xsens MTi-700 & 400 & - \\
            INS & Applanix LVX & 100 & - \\
            \hline\hline
        \end{tabular} }
    \begin{tablenotes}
    \footnotesize
    \item $^1$ resolution, $^2$ number of channels.
    \end{tablenotes}
\vspace{-0.4cm}
\end{table}

\begin{figure}[!h]
  \centering
  \setlength{\belowcaptionskip}{-0.3cm}
  \includegraphics[width=0.46\textwidth]{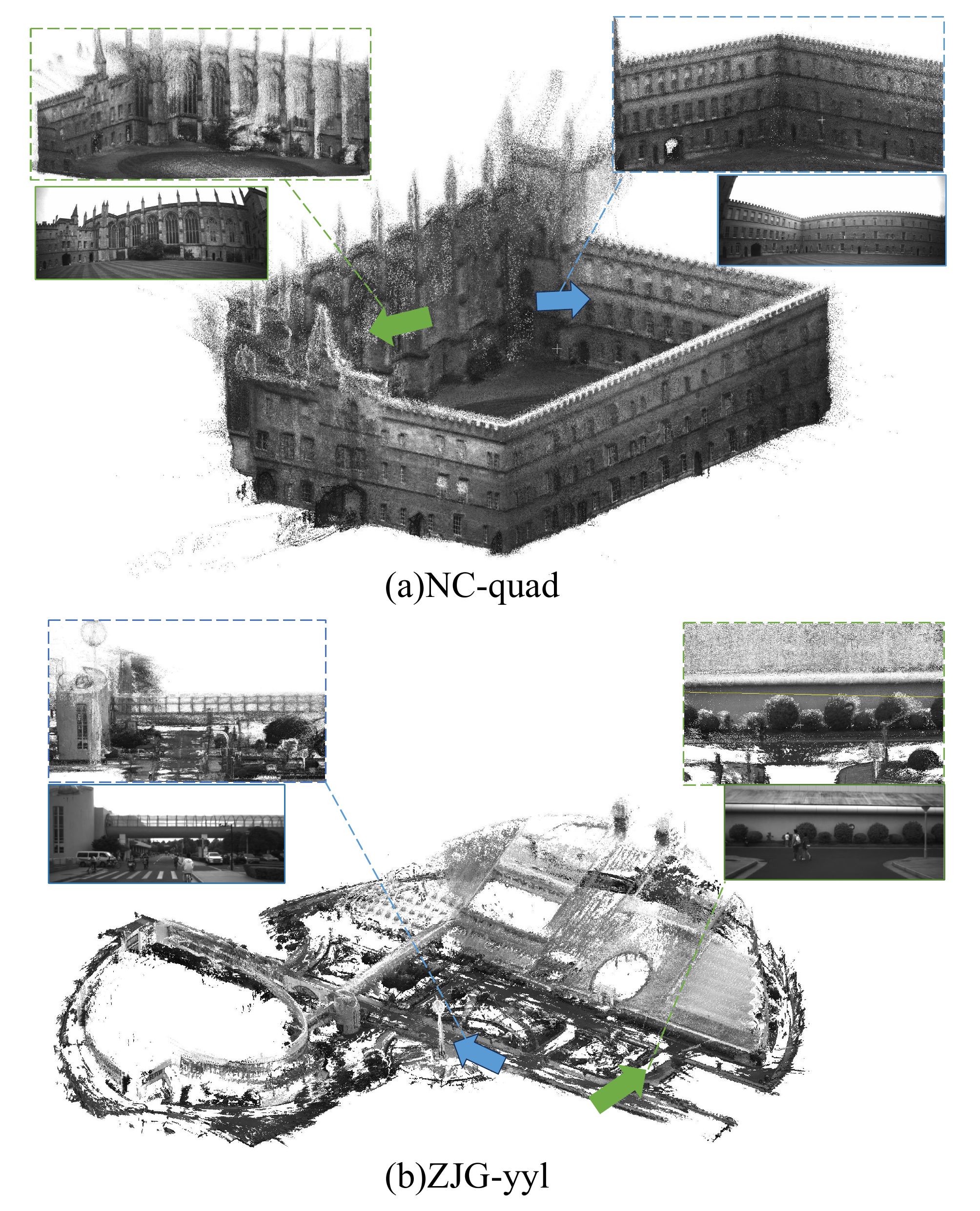}
  \caption{Dense reconstruction results using our method. Different colored arrows represent various directions of observation. The dashed box shows the result of the dense reconstruction as viewed from the arrow of the corresponding color, and the solid box shows the corresponding real-world scene.}
  \label{dense_reconstruction}
\vspace{-0.2cm}
\end{figure}

\subsection{MAPPING ACCURACY AND ROBUSTNESS}\label{sec:mapping_eval}

\textcolor{black}{To comprehensively evaluate our visual mapping pipeline, we design three complementary experiments that address fundamental aspects of mapping performance. First, we assess the baseline accuracy of our visual mapping approach against established methods to demonstrate its reconstruction quality. Second, we investigate how pose prior information from our multi-camera VINS affects initial map construction, as reliable pose estimates are crucial for robust feature matching in challenging environments. Third, we examine the benefits of integrating external sensor data for scale consistency, particularly relevant for large-scale field deployments where visual-inertial methods may struggle with drift accumulation. These experiments collectively validate both the core visual mapping capabilities and the system's adaptability to different sensor configurations.}

\begin{table}[tp]
\centering
  \caption{Mapping accuracy comparison in ZJG and NC datasets \textcolor{black}{(ATE position RMSE in meters). All methods evaluated without loop closure to assess VIO trajectory quality.}}\label{accuracy_compare}
\begin{tabular}{ccccc}
\hline \hline
\multirow{2}{*}{Algorithm} & \multicolumn{2}{l}{ZJG-qsdjt-0924} & \multicolumn{2}{l}{NC-quad-easy} \\
                           & mean            & std           & mean           & std            \\
\hline
OpenVINS                   & 14.05           & 8.17          & 1.59           & 0.59           \\
Maplab                     & 20.64           & 6.69          & 1.25           & 0.43           \\
VINS-Fusion                & 1.98            & \textbf{0.37}          & 0.45           & 0.14           \\
Ours                       & \textbf{1.55}            & 0.61          & \textbf{0.31}   & \textbf{0.12}  \\
\hline \hline
\end{tabular}
\label{tab:addlabel}
\vspace{-0.2cm}
\end{table}

\begin{figure}[tp]
    \centering
    \includegraphics[width=0.48\textwidth]{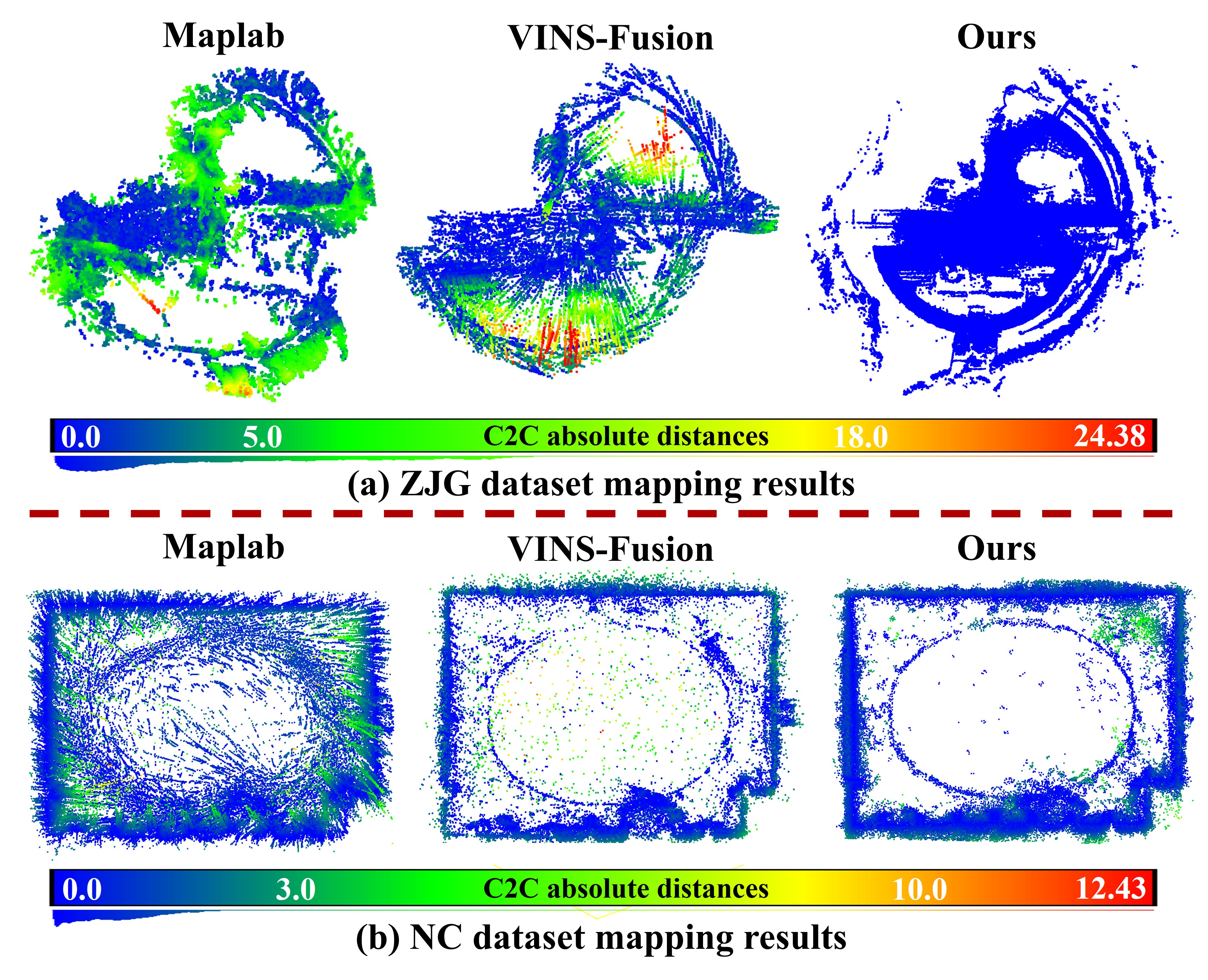}
    \caption{Map accuracy evaluation using ${E}_{mapping}^p$. Point clouds are color-coded by reconstruction error for different algorithms, with blue indicating low error and red indicating high error.}\label{fig:accuracy-test}
\vspace{-0.4cm}
\end{figure}

\begin{figure}[tp]
    \centering
    \includegraphics[width=0.48\textwidth]{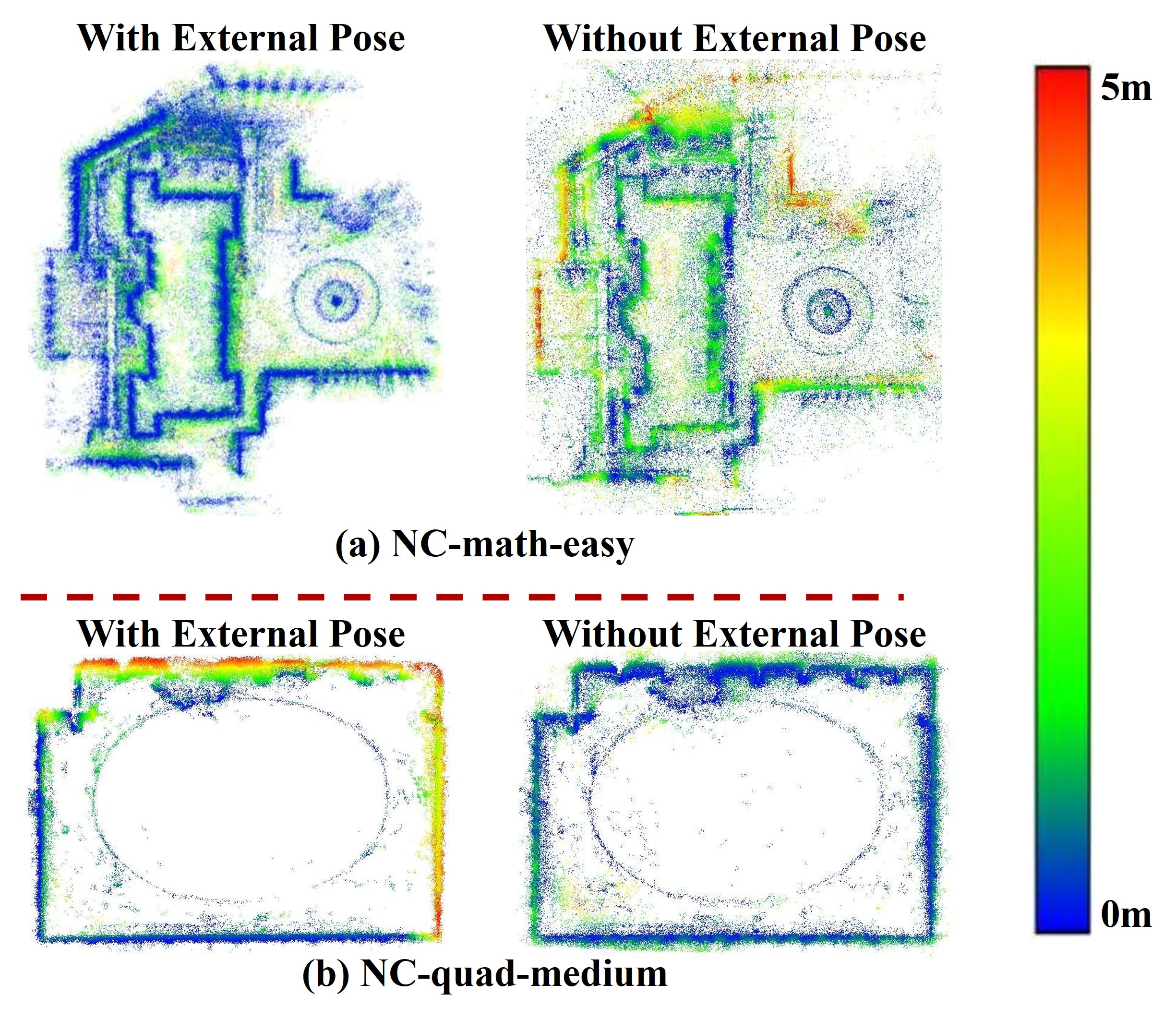}
    \caption{Mapping accuracy comparison results of whether or not to use external pose. }\label{fig:lidar-visual-comparison}
\vspace{-0.4cm}
\end{figure}

\subsubsection{Accuracy verification}\label{sec:map-exp2}
\textcolor{black}{To establish baseline mapping performance, we compare our reconstruction accuracy against established open-source systems with mapping capabilities on both the NC and ZJG datasets. We employ two evaluation approaches: quantitative point-level error analysis to measure geometric precision, and qualitative dense reconstruction assessment to evaluate visual completeness and structural fidelity.}

\textbf{Quantitative comparison.} The ground truth map for the NC and ZJG dataset are obtained by Leica BLK360, a survey-grade 3D imaging laser scanner, and results obtained from the LiDAR-inertial odometry algorithm~\cite{fastlio2}. After removing some outlier points by statistical outlier filter, we evaluate reconstruction results of each algorithm with the ground truth using ${E}_{mapping}^p$ metric in (\ref{eq:map_pt}) and visualize the matching error for each point, as depicted in Fig.\ref{fig:accuracy-test}. Table~\ref{tab:addlabel} presents the quantitative comparison across all methods, showing that our approach achieves 1.55m mean error on ZJG-qsdjt-0924 and 0.31m on NC-quad-easy, outperforming OpenVINS, Maplab, and VINS-Fusion. The comparison highlights that our method achieves higher reconstruction accuracy than both Maplab and VINS-Fusion.

\textbf{Qualitative comparison.} To further evaluate the reconstruction accuracy visually, we perform a dense reconstruction based on the sparse results obtained from the earlier quantitative comparison. As illustrated in Fig.\ref{dense_reconstruction}, the dense maps generated by our method closely match the actual camera-captured images, and the objects within these images have well-defined contours. This demonstrates the effectiveness of our dense reconstruction approach.


\vspace{-0.2cm}
\subsubsection{Impact of VINS Pose Prior on Mapping}

\textcolor{black}{To assess the impact of pose priors on offline map construction, we conduct experiments on the qsdjt scene from the ZJG dataset. We compare three approaches for providing initial pose information to the offline STAGE I reconstruction: pose-independent sequential matching without prior information, multi-camera VINS priors without loop closure corrections, and loop closure-enhanced multi-camera VINS priors. Each approach uses spatial matching to pair images based on the respective pose priors for subsequent initial reconstruction.}

\begin{figure*}[tbh]
  \centering
  \includegraphics[width=1\hsize]{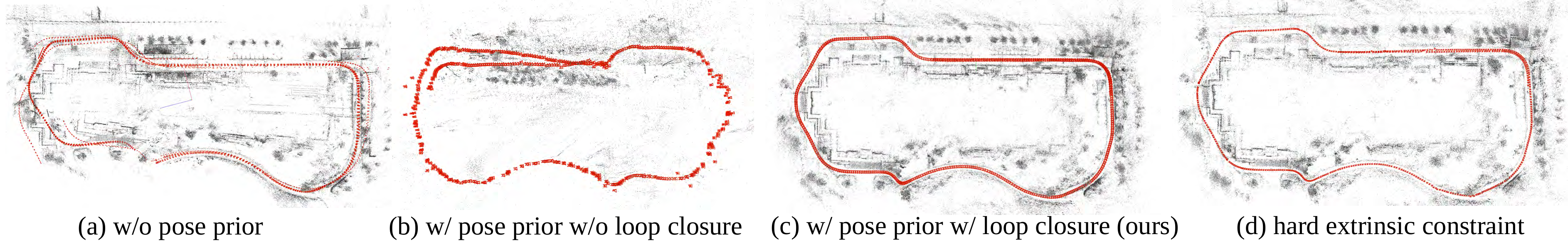}
  \caption{Initial map reconstruction results demonstrating the impact of pose priors. (a) Initial map generated without using pose prior (sequential matching). (b) Initial map using VINS pose prior without loop closure. (c) Initial map using VINS pose prior with loop closure (our STAGE I approach), resulting in a clearer reconstruction. (d) Result after refining poses using extrinsic parameters as hard constraints, showing increased noise.}
  \label{fig:init_compare}
\vspace{-0.2cm} 
\end{figure*}

\textcolor{black}{Qualitative results appear in Fig.\ref{fig:init_compare}. Without pose priors (Fig.\ref{fig:init_compare}a), the initial map exhibits fragmentation and reduced accuracy due to unreliable correspondences in complex scenes. VINS priors without loop closure (Fig.\ref{fig:init_compare}b) offer limited improvement by constraining the feature matching search space. In contrast, loop closure-enhanced priors (Fig.\ref{fig:init_compare}c) substantially improve matching robustness and reconstruction clarity, yielding coherent 3D structure.}

\textcolor{black}{We further examine extrinsic parameter treatment during STAGE II (Sec.~\ref{sec:offline_map}) refinement, comparing our soft constraint optimization against hard constraint enforcement. Fig.\ref{fig:init_compare}d shows that hard constraints introduce noise and degrade map quality relative to our approach (Fig.\ref{fig:init_compare}c). This validates our strategy: leveraging loop closure-enhanced VINS priors for initialization, then optimizing extrinsics as soft constraints to accommodate calibration uncertainties.}


\vspace{-0.2cm}
\subsubsection{Impact of External LiDAR Pose Integration}\label{sec:map-exp3} 

To evaluate the benefits of incorporating external pose information in our visual mapping pipeline, we conduct comparative experiments in two scenarios of the NC dataset: math-easy (a large-scale environment) and quad-medium (a smaller space). We use FAST-LIO~\cite{xu2021fast} to process LiDAR data and generate external pose observations, which are then integrated into our mapping framework as described in Sec.~\ref{sec:mapping}.

The mapping accuracy comparison results shown in Fig.\ref{fig:lidar-visual-comparison} demonstrate that incorporating external pose observations significantly improves mapping accuracy in large-scale environments (NC-math-easy), while providing marginal benefits in smaller spaces (quad-medium). This confirms our hypothesis that absolute pose references are particularly valuable for maintaining scale consistency in extensive areas.

We further evaluated localization performance using maps constructed with and without external pose information. As shown in Table~\ref{table:lidar_visual_map}, localization accuracy improves substantially when using externally-augmented maps in large environments (NC-math), while the improvement is minimal in smaller spaces (NC-quad). These results highlight the value of sensor fusion approaches in challenging large-scale field deployments, where single-modality methods often struggle with drift and scale inconsistency.


\begin{table}[]
\caption{Translation part (m) of local-based trajectory error across maps.}\label{table:lidar_visual_map}
\resizebox{0.47\textwidth}{!}{
\begin{tabular}{lcccc}
\hline \hline
               & \multicolumn{2}{c}{With external pose} & \multicolumn{2}{c}{Without external pose} \\
               & mean               & std               & mean                & std                 \\ \hline
NC-math-medium & 1.01               & 0.44              & 1.69                & 0.86                \\
NC-math-hard   & 3.76               & 1.75              & 4.12                & 2.01   \\ 
NC-quad-easy & 0.91               & 0.39              &       0.96           & 0.49        \\ 
NC-quad-hard   & 1.14               & 0.44              &      1.15    &   0.45               
\\ \hline \hline
\end{tabular}}
\vspace{-0.4cm}
\end{table}

\vspace{-0.2cm}
\textcolor{black}{\subsubsection{Discussion and Practical Implications}}\label{sec:map-discussion}

\textcolor{black}{The mapping experiments reveal fundamental relationships between environment scale, constraint topology, and reconstruction quality that inform principled system design.}

\textcolor{black}{External pose sensors exhibit scale-dependent effectiveness because visual-inertial drift accumulates proportionally to trajectory length. Table~\ref{table:lidar_visual_map} demonstrates 40\% error reduction in large environments versus 5\% in compact spaces, indicating that absolute scale references become valuable only when drift dominates the error budget. This suggests sensor selection should be driven by trajectory characteristics rather than environment type alone.}

\textcolor{black}{The pose prior ablation in Fig.\ref{fig:init_compare} illustrates that constraint graph topology critically affects reconstruction coherence. Loop closures transform sequential constraints into cyclic graphs that distribute accumulated errors, preventing fragmentation. Similarly, soft extrinsic constraints outperform hard constraints by accommodating calibration uncertainties rather than forcing infeasible geometric relationships. These findings indicate that trajectory planning should prioritize loop structures and optimization formulations should embrace uncertainty modeling.}

\textcolor{black}{For practical deployment, trajectories under 100 meters achieve sufficient accuracy with visual-inertial mapping alone, while trajectories exceeding 500 meters justify external sensors despite increased complexity. Intermediate scales require application-specific trade-offs between accuracy requirements and system constraints, though dynamic environments and loop closure requirements may further constrain these design choices.}


\vspace{0.2cm}
\subsection{MATCHING ACCURACY AND ROBUSTNESS}\label{sec:matching_eval}

\textcolor{black}{Robust feature matching is essential for reliable localization under long-term environmental changes. This section validates two critical components of our localization pipeline: the deterministic initialization module (Sec.~\ref{sec:init}) that establishes initial alignment with guaranteed convergence, and the online matching module (Sec.~\ref{sec:online_match}) that provides continuous map observations despite extreme outlier rates. We evaluate these components through complementary experiments that stress-test their robustness against seasonal appearance changes, illumination variations, and substantial viewpoint differences.}

\textcolor{black}{Our evaluation leverages three datasets with distinct characteristics. The YQ dataset provides controlled seasonal variation with summer mapping sessions and winter localization attempts. The NC dataset features circular trajectories that induce large viewpoint changes despite compact spatial extent. The ZJG dataset presents extended outdoor scenarios with ongoing construction, vegetation growth, and varying weather conditions. Together, these datasets enable systematic assessment of matching robustness across the operational envelope of field robotics applications.}

\vspace{-0.2cm}
\subsubsection{Deterministic Initialization}\label{sec:init_eval}

Robust initialization is mission-critical for localization systems, as failure at this stage prevents subsequent operation entirely. \textcolor{black}{We compare our deterministic convergence method against RANSAC-based probabilistic approaches on six representative cases from the YQ dataset. For each selected case, we execute each method 100 times to gather statistical results, which include the standard deviation, translation error, rotation error, and computation time. Note that the translation and rotation error are computed using the alignment evaluation metric in (\ref{eq:alignt}) and (\ref{eq:alignr}). The baseline 2P RANSAC uses 100 iterations, while 2P10K employs 10,000 iterations to demonstrate the limits of probabilistic convergence.}

\textcolor{black}{Table~\ref{table.repeat} reveals fundamental differences in convergence behavior among IMU-aided pose estimation methods. All compared approaches exploit gravity alignment from inertial measurements to reduce the problem from 6DoF to 4DoF. Standard RANSAC exhibits high variance across trials, with translation standard deviations ranging from 0.033m to 0.554m depending on inlier rate and feature count. Increasing iterations to 10,000 substantially improves repeatability, reducing standard deviation to near-zero levels, but requires 4-6 seconds per initialization, which is impractical for time-sensitive field operations. Our deterministic method achieves zero variance across all trials while maintaining competitive accuracy and reasonable computation time (0.1-0.7 seconds). This guaranteed convergence stems from exhaustive search over the decoupled rotation and translation spaces, ensuring global optimality without sampling uncertainty.}

\begin{table}[tbp]
\scriptsize
\caption{Deterministic initialization performance comparison using IMU-aided minimal solvers across six scenarios from YQ dataset.}
\begin{center}
\resizebox{0.5\textwidth}{!}{
\begin{threeparttable}
\renewcommand\arraystretch{1.2}
\begin{tabular}{l c c c c c c c}
\hline
{ExpID} & {$N/\omega$} & {Method} & \makecell[c]{$\triangle T$ \\ (m)} & \makecell[c]{$\sigma T$ \\ (m)} & \makecell[c]{$\triangle R$ \\ (\degree)} & \makecell[c]{$\sigma R$ \\ (\degree)} & \makecell[c]{Time \\ (s)}\\
\hline
\multirow{3}{*}{01} & \multirow{3}{*}{26~/~0.82} & {2P} & \textcolor{black}{0.773} & \textcolor{black}{0.474} & \textcolor{black}{1.644} & \textcolor{black}{0.633} & {0.074} \\
& & {2P10K} & \textcolor{black}{0.293} & \textcolor{black}{0.002} & \textcolor{black}{0.892} & \textcolor{black}{0.001} & \textcolor{black}{4.324} \\
& & {Ours} & \textbf{0.154} & \textbf{0.000} & \textbf{0.498} & \textbf{0.000} & {0.100} \\
\cline{3-8}
\multirow{3}{*}{02} & \multirow{3}{*}{68~/~0.78} & {2P} & \textcolor{black}{0.243} & \textcolor{black}{0.033} & \textcolor{black}{0.295} & \textcolor{black}{0.041} & {0.080} \\
& & {2P10K} & \textcolor{black}{0.263} & \textcolor{black}{0.000} & \textcolor{black}{0.269} & \textcolor{black}{0.000} & \textcolor{black}{6.187} \\
& & {Ours} & \textbf{0.049} & \textbf{0.000} & \textbf{0.267} & \textbf{0.000} & {0.320} \\
\cline{3-8}
\multirow{3}{*}{03} & \multirow{3}{*}{15~/~0.47} & {2P} & \textcolor{black}{0.583} & \textcolor{black}{0.554} & \textcolor{black}{1.422} & \textcolor{black}{1.725} & {0.064} \\
& & {2P10K} & \textcolor{black}{0.205} & \textcolor{black}{0.000} & \textcolor{black}{0.251} & \textcolor{black}{0.000} & \textcolor{black}{6.325} \\
& & {Ours} & \textbf{0.187} & \textbf{0.000} & \textbf{0.130} & \textbf{0.000} & {0.080} \\
\cline{3-8}
\multirow{3}{*}{04} & \multirow{3}{*}{49~/~0.65} & {2P} & \textcolor{black}{0.397} & \textcolor{black}{0.208} & \textcolor{black}{0.891} & \textcolor{black}{0.549} & {0.071} \\
& & {2P10K} & \textcolor{black}{0.335} & \textcolor{black}{0.003} & \textcolor{black}{0.785} & \textcolor{black}{0.011} & \textcolor{black}{4.855} \\
& & {Ours} & \textbf{0.283} & \textbf{0.000} & \textbf{0.596} & \textbf{0.000} & {0.150} \\
\cline{3-8}
\multirow{3}{*}{05} & \multirow{3}{*}{82~/~0.72} & {2P} & \textcolor{black}{0.106} & \textcolor{black}{0.077} & \textcolor{black}{0.335} & \textcolor{black}{0.102} & {0.084} \\
& & {2P10K} & \textcolor{black}{0.095} & \textcolor{black}{0.020} & \textcolor{black}{0.390} & \textcolor{black}{0.024} & \textcolor{black}{5.954} \\
& & {Ours} & \textbf{0.045} & \textbf{0.000} & \textbf{0.320} & \textbf{0.000} & {0.420} \\
\cline{3-8}
\multirow{3}{*}{06} & \multirow{3}{*}{138~/~0.38} & {2P} & \textcolor{black}{0.205} & \textcolor{black}{0.364} & \textcolor{black}{0.478} & \textcolor{black}{0.842} & {0.080} \\
& & {2P10K} & \textcolor{black}{0.210} & \textcolor{black}{0.151} & \textcolor{black}{0.468} & \textcolor{black}{0.033} & \textcolor{black}{6.097} \\
& & {Ours} & \textbf{0.125} & \textbf{0.000} & \textbf{0.390} & \textbf{0.000} & {0.680} \\
\hline
\end{tabular}
\begin{tablenotes}
	\footnotesize
	\item[1] $N$ denotes the number of features, $\omega$ is the inlier rate and $\sigma$ is the standard deviation.
\end{tablenotes}
\renewcommand\arraystretch{1.0}
\end{threeparttable}}
\end{center}
\label{table.repeat}
\vspace{-0.4cm}
\end{table}

\begin{figure}[t]
  \centering
  \includegraphics[width=0.48\textwidth]{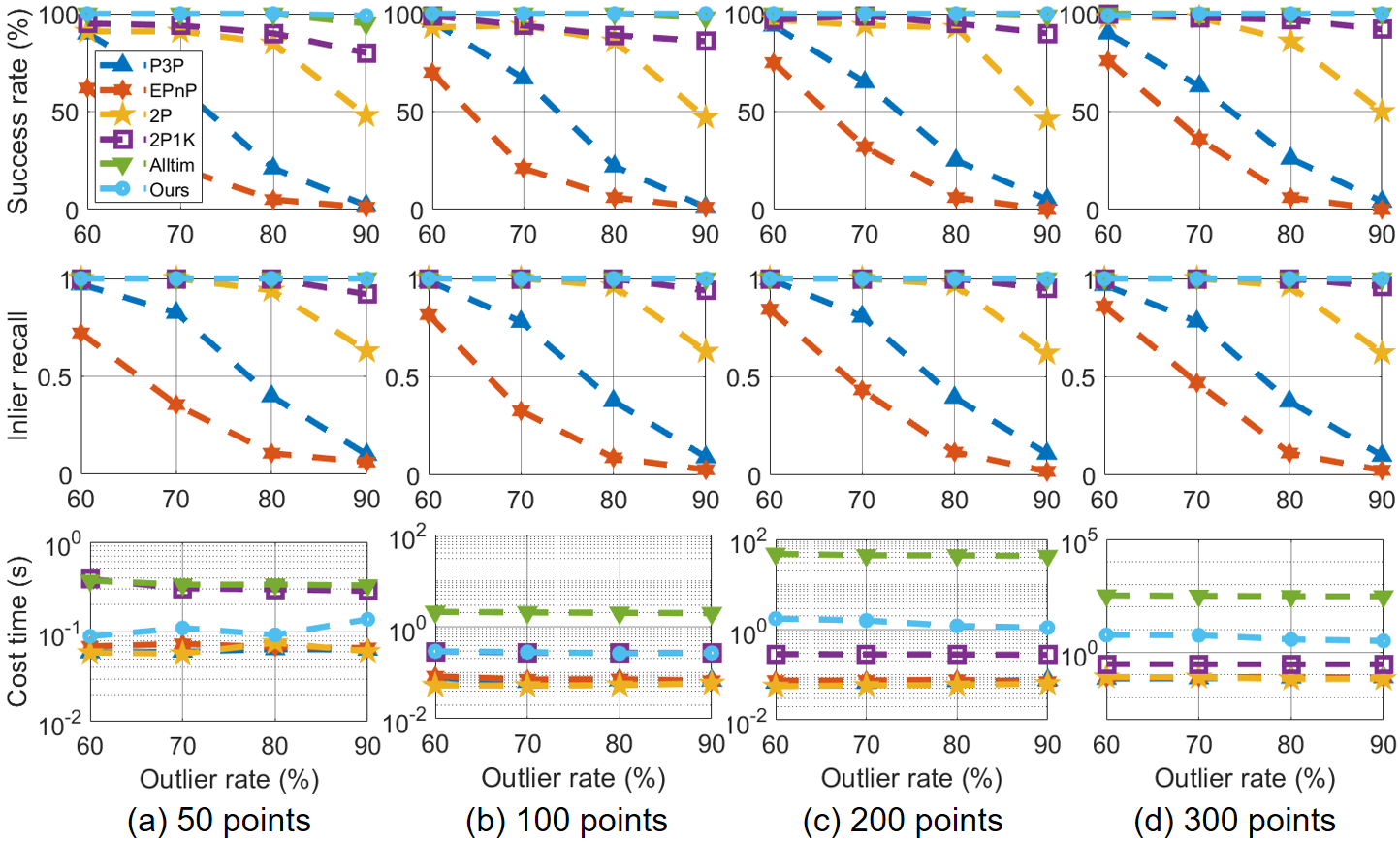}
  \caption{\textcolor{black}{Performance comparison under varying outlier rates (60-90\%) and correspondence counts.}}
  \label{fig:robustness-outlier}
  \vspace{-0.4cm}
\end{figure}

\textcolor{black}{To further validate robustness under controlled conditions, we conduct synthetic experiments with varying outlier rates and feature counts. Fig.~\ref{fig:robustness-outlier} compares our deterministic method against RANSAC-based approaches (P3P, EPnP, 2P, 2P1K) and an exhaustive baseline (Alltim) across three metrics: success rate (0.05m translation, 0.5° rotation threshold), inlier recall, and computation time. Both Alltim and our method achieve 100\% inlier recall, outperforming all RANSAC variants. While 2P1K improves over P3P/EPnP (fewer DoFs) and 2P (more iterations), it cannot guarantee deterministic convergence. Our method bridges the efficiency-determinism gap: maintaining 100\% inlier recall while achieving higher efficiency than Alltim through time complexity proportional to correspondence count. For typical vehicle localization (<200 correspondences), our method completes within seconds.}

\begin{table*}[thp]
\caption{Success rate comparison of mono-camera and multi-camera algorithms using matching error metric under different thresholds.}
\begin{center}
\resizebox{1.0\textwidth}{!}{
\begin{tabular}{cl c c ccc}
\hline\hline
  \multirow{3}{*}{Task}&session& {YQ-2017-0827} & {YQ-2017-0828} & {YQ-2018-0129}  & NC-quad-hard & NC-math-hard\\
 &\makecell[l]{m \\deg} & \makecell[c]{0.25 / 0.5 / 1.0 / 5.0 \\ 2.0 / 5.0 / 8.0 / 10.0}& \makecell[c]{0.25 / 0.5 / 1.0 / 5.0 \\ 2.0 / 5.0 / 8.0 / 10.0}& \makecell[c]{0.25 / 0.5 / 1.0 / 5.0 \\ 2.0 / 5.0 / 8.0 / 10.0}& \makecell[c]{0.25 / 0.5 / 1.0 / 5.0 \\ 2.0 / 5.0 / 8.0 / 10.0}&\makecell[c]{0.25 / 0.5 / 1.0 / 5.0 \\ 2.0 / 5.0 / 8.0 / 10.0}\\
\hline
 \multirow{4}{*}{Mono}&P3P & \textcolor{black}{15.6} / \textcolor{black}{28.6} / \textcolor{black}{38.2} / \textcolor{black}{49.4}  &
\textcolor{black}{16.2} / \textcolor{black}{32.1} / \textcolor{black}{39.1} / \textcolor{black}{47.1}  & \textcolor{white}{0}{9.2} / \textcolor{black}{18.3} / \textcolor{black}{27.8} / \textcolor{black}{45.2}  & \textcolor{black}{2.75} / {10.56} / \textcolor{black}{26.68} / \textcolor{black}{55.13}&\textcolor{black}{2.05} / \textcolor{black}{10.36} / \textcolor{black}{30.64} / \textcolor{black}{74.83}\\
 &EPnP & \textcolor{black}{15.9} / \textcolor{black}{29.2} / \textcolor{black}{38.6} / \textcolor{black}{50.7} & \textcolor{black}{17.8} / \textcolor{black}{33.0} / \textcolor{black}{\textbf{39.6}} / \textcolor{black}{\textbf{47.6}} & \textcolor{white}{0}{9.8} / \textcolor{black}{20.7} / \textcolor{black}{28.8} / \textcolor{black}{45.6}  & \textcolor{black}{2.66} / {10.96} / \textcolor{black}{28.50} / \textcolor{black}{54.86}&\textcolor{black}{2.43} / \textcolor{black}{10.74} / \textcolor{black}{32.35} / \textcolor{black}{74.21}\\
 &UPnP& \textcolor{black}{19.2} / \textcolor{black}{33.1} / \textcolor{black}{39.2} / \textcolor{black}{51.8} & \textcolor{black}{19.6} / \textcolor{black}{\textbf{34.3}} / \textcolor{black}{38.2} / \textcolor{black}{47.4} & \textcolor{black}{10.4} / \textcolor{black}{21.3} / \textcolor{black}{28.2} / \textcolor{black}{46.8}  & \textcolor{black}{2.62} / {10.52} / \textcolor{black}{27.08} / \textcolor{black}{54.86}&\textcolor{black}{1.95} / \textcolor{black}{10.50} / \textcolor{black}{30.71} / \textcolor{black}{74.66}\\
 &2P & \textbf{\textcolor{black}{21.9} / \textcolor{black}{33.8} / \textcolor{black}{41.0} / \textcolor{black}{54.6}}& \textcolor{black}{\textbf{20.9}} / \textcolor{black}{31.9} / \textcolor{black}{37.1} / \textcolor{black}{44.3} & \textbf{\textcolor{black}{13.9} / \textcolor{black}{24.4} / \textcolor{black}{33.5} / \textcolor{black}{53.5}}  & \textbf{\textcolor{black}{8.17} / \textcolor{black}{21.66} / \textcolor{black}{39.37} / \textcolor{black}{57.08}}&\textbf{\textcolor{black}{5.51} / \textcolor{black}{19.66} / \textcolor{black}{44.63} / \textcolor{black}{80.03}}\\
 \hline
 \multirow{3}{*}{Multi}&GP3P & \textcolor{black}{22.9} / \textcolor{black}{36.5} / \textcolor{black}{42.8} / \textcolor{black}{54.3}  &
\textcolor{black}{22.6} / \textbf{\textcolor{black}{36.8}} / \textcolor{black}{42.7} / \textbf{\textcolor{black}{52.2}}  & \textcolor{black}{15.2} / \textcolor{black}{22.1} / \textcolor{black}{31.0} / \textcolor{black}{53.5}  & {10.46} / {26.95} / \textcolor{black}{45.74} / \textcolor{black}{60.46}&\textcolor{black}{7.66} / \textcolor{black}{24.08} / \textcolor{black}{52.80} / \textcolor{black}{88.78}\\
 &GPnP& \textcolor{black}{21.1} / \textcolor{black}{35.5} / \textcolor{black}{41.4} / \textcolor{black}{52.4} & \textcolor{black}{22.9} / \textcolor{black}{34.8} / \textcolor{black}{41.3} / \textcolor{black}{50.2} & \textcolor{black}{14.3} / \textcolor{black}{21.8} / \textcolor{black}{30.2} / \textcolor{black}{52.0}  & {10.11} / {23.40} / \textcolor{black}{41.13} / \textbf{\textcolor{black}{61.17}}&\textcolor{black}{6.97} / \textcolor{black}{25.99} / \textcolor{black}{54.45} / \textcolor{black}{89.33}\\
 &MC2P & \textbf{\textcolor{black}{23.0} / \textcolor{black}{37.1} / \textcolor{black}{45.6} / \textcolor{black}{55.8}} & \textbf{\textcolor{black}{23.9}} / \textcolor{black}{36.6} / \textbf{\textcolor{black}{43.2}} / \textcolor{black}{51.7} & \textbf{\textcolor{black}{15.3} / \textcolor{black}{26.4} / \textcolor{black}{37.2} / \textcolor{black}{57.6}}  & \textbf{\textcolor{black}{12.41} / {27.48} / \textcolor{black}{46.81} / \textcolor{black}{61.17}}&\textbf{\textcolor{black}{9.71} / \textcolor{black}{26.27} / \textcolor{black}{55.13} / \textcolor{black}{90.42}}\\
\hline\hline
\end{tabular}}
\end{center}
\label{table.whole}
\vspace{-0.5cm}
\end{table*}

\begin{figure}[tbp]
    \centering
    \includegraphics[width=0.5\textwidth]{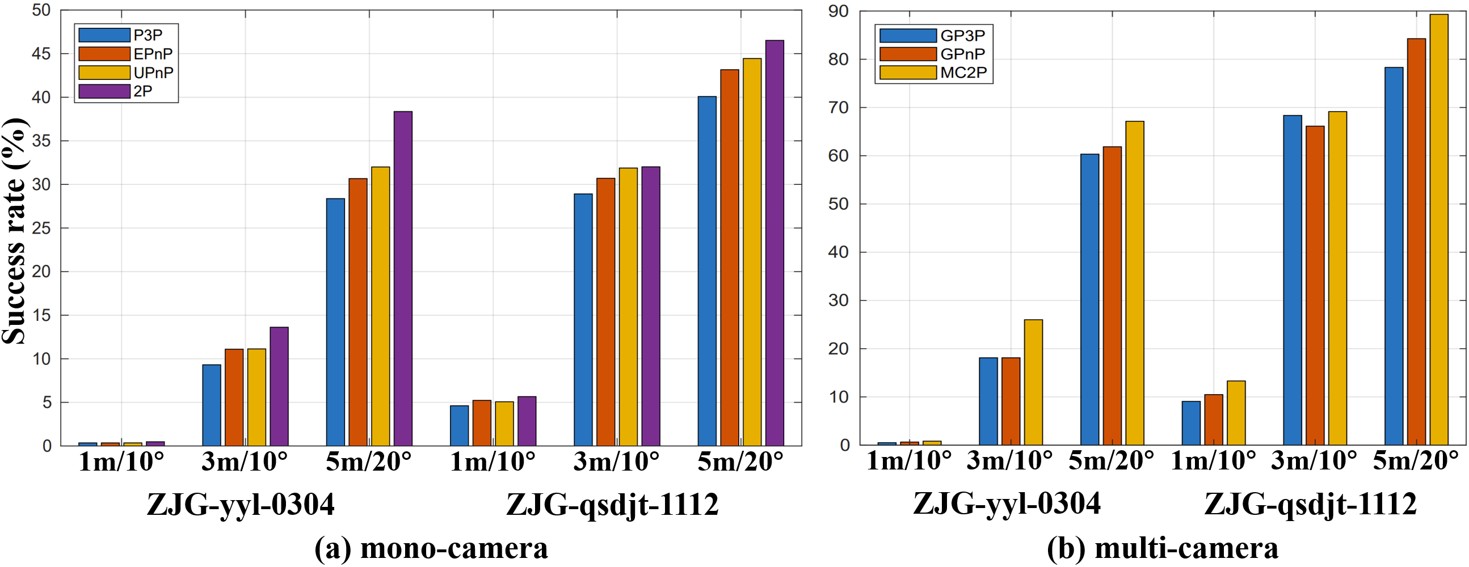}
   \caption{Success rate comparison on ZJG dataset using matching error metric under different thresholds.}
    \label{fig.ZJG_suc}
\vspace{-0.5cm}
\end{figure}

\vspace{-0.4cm}
\textcolor{black}{\subsubsection{Online Matching Robustness}}\label{sec:online_match_eval}

\textcolor{black}{After initialization, continuous localization requires robust online matching between query images and map keyframes. We evaluate outlier rejection performance using NetVLAD~\cite{NetVLAD} for image retrieval, SuperPoint~\cite{detone2018superpoint} and LightGlue~\cite{lightglue} for feature extraction and matching. For monocular pose estimation, we aggregate feature matches from the top-5 retrieved map images to construct 2D-3D correspondence sets. The RANSAC framework employs different minimal solutions for model estimation, including the proposed 2P, and established methods P3P~\cite{gao2003complete}, EPnP~\cite{lepetit2009EPnP}, and UPnP~\cite{kneip2014UPnP}. For multi-camera pose estimation, matches from all cameras are aggregated independently before joint estimation using MC2P, GP3P~\cite{kneip2013using}, and GPnP~\cite{kneip2013using}. We compute pose estimation errors using the alignment metrics in (\ref{eq:alignt}) and (\ref{eq:alignr}), then evaluate robustness by counting success rates at varying accuracy thresholds.}

\textcolor{black}{Table~\ref{table.whole} and Fig.\ref{fig.ZJG_suc} presents the success rate comparison across YQ, NC, and ZJG datasets. Multi-camera algorithms consistently outperform monocular approaches, with MC2P achieving 4.5$\times$ higher success rate than P3P at the 0.25m threshold on NC-quad-hard. The multi-camera advantage is particularly pronounced on the NC dataset where surround-view cameras capture complementary structural information from circular trajectories. Among monocular methods, the proposed 2P solution demonstrates superior robustness in challenging scenarios. On both NC-quad-hard and NC-math-hard, 2P achieves 2-3$\times$ higher success rates than traditional three-point methods at the 0.25m threshold. On YQ-2018-0129 with seasonal variations, 2P achieves 13.9\% success compared to 9.2-10.4\% for three-point methods. Fig.\ref{fig.ZJG_suc} confirms this robustness advantage extends to the ZJG dataset. Conversely, in favorable conditions such as YQ-2017-0828 with similar mapping and localization lighting, 6DoF methods achieve marginally better accuracy than the proposed 4DoF approach when feature matching quality is high.}

\begin{figure*}[tbp]
    \centering
    \includegraphics[width=0.96\textwidth]{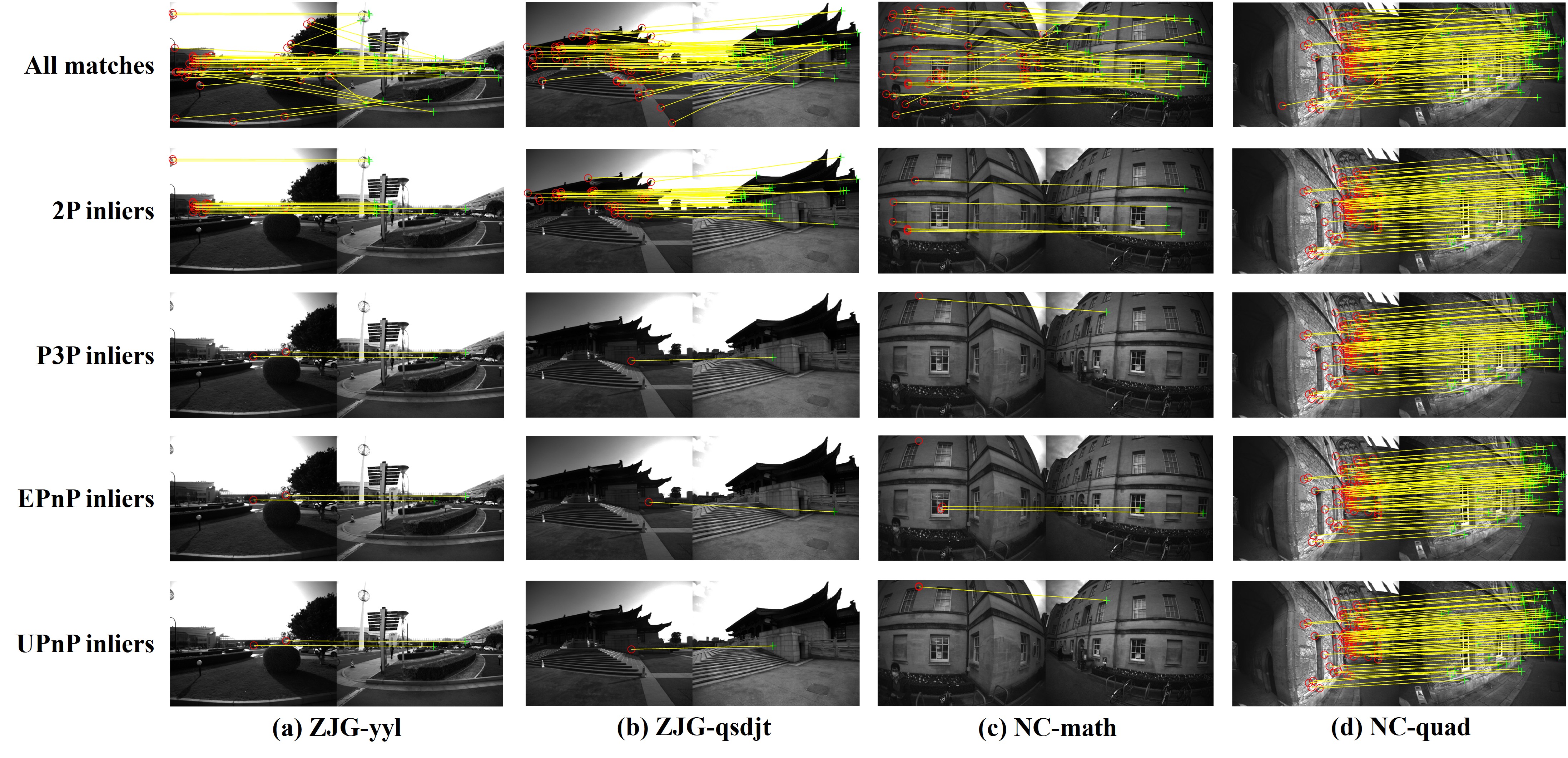}
    \caption{Visualization of matching inliers identified by different methods. \textcolor{black}{Comparison of 2P, P3P, EPnP, and UPnP results on ZJG and NC datasets, where the 2P method yields more correct inlier correspondences than the others.}}
    \label{fig.matchcases}
\vspace{-0.4cm}
\end{figure*}

\textcolor{black}{Fig.\ref{fig.matchcases} visualizes matching results for representative cases. ZJG-yyl (Fig.\ref{fig.matchcases}a) exemplifies a large-scale unstructured scene with sparse features and substantial viewpoint changes, where 2P successfully filters correct correspondences while P3P, EPnP, and UPnP fail entirely. Similar patterns appear in ZJG-qsdjt and NC-math (Fig.\ref{fig.matchcases}b-c) with extreme viewpoint variations. In contrast, NC-quad (Fig.\ref{fig.matchcases}d) with abundant structural features and moderate viewpoint differences enables all methods to achieve comparable outlier rejection performance.}

\vspace{-0.2cm}
\textcolor{black}{\subsubsection{Discussion and Practical Implications}}\label{sec:matching_discussion}

\textcolor{black}{The matching experiments reveal fundamental trade-offs between algorithmic guarantees and computational efficiency. Deterministic initialization provides complete convergence guarantees through exhaustive search over decoupled solution spaces, eliminating variance in pose estimation outcomes. This determinism comes at computational cost but addresses a critical deployment constraint: initialization failure terminates localization capability entirely, whereas modest computational overhead can be tolerated during system startup. For online matching, reduced minimal sample complexity directly translates to improved RANSAC convergence probability under high outlier rates. The 4DoF formulation leveraging IMU gravity alignment reduces required correspondences from three or four to two, fundamentally altering the sampling statistics in outlier-dominated scenarios. However, this robustness gain involves an inherent accuracy trade-off when operating in feature-rich environments with precise calibration, where the additional degrees of freedom in 6DoF formulations can be reliably estimated.}

\textcolor{black}{These findings inform deployment-specific algorithm selection strategies. Deterministic methods should be prioritized for mission-critical applications despite increased startup time. For continuous operation, the choice between 4DoF and 6DoF formulations should be driven by environmental characteristics and sensor quality. Unstructured outdoor environments subject to long-term appearance variations benefit from IMU-aided minimal formulations. Conversely, controlled indoor environments with stable lighting and accurate multi-camera calibration may achieve superior performance through 6DoF optimization. Algorithm selection should reflect operational requirements and expected data quality rather than environment type alone.}

\begin{figure*}[!tb]
  \centering
  \setlength{\belowcaptionskip}{-0.3cm}
  \includegraphics[width=0.96\hsize]{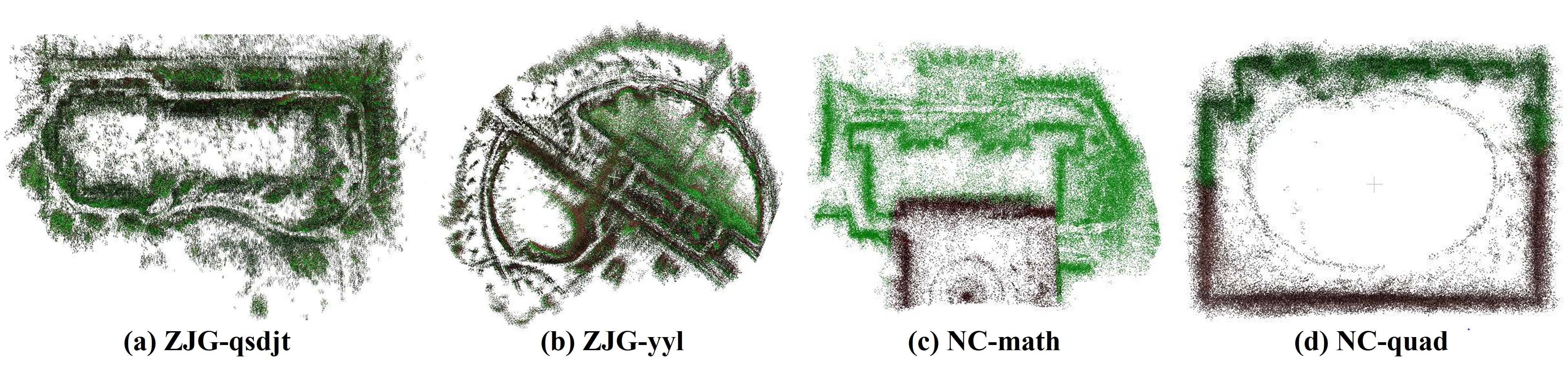}
  \caption{Multi-map experiment result, which shows the merged result of the final estimated extrinsic by the proposed system}
  \label{fig:multimap_align}
\vspace{-0.4cm}
\end{figure*}

\vspace{0.4cm}
\subsection{LOCALIZATION ACCURACY AND ROBUSTNESS}\label{sec:localization_eval}
\textcolor{black}{Our localization module integrates observations from multiple isolated maps and multi-camera systems to provide robust state estimation. This section validates two key capabilities: whether fusing multiple maps improves accuracy compared to single-map localization, and how different camera configurations affect performance across diverse environments. We evaluate these aspects on the NC and ZJG datasets, which offer complementary challenges. NC features compact spaces with large viewpoint variations due to circular trajectories, while ZJG presents large-scale outdoor scenarios with long-term appearance changes. Performance is measured using local-based trajectory error ($E^{local}_{trajectory}$, Eq.~(\ref{eq.local_traj})) for causal tracking accuracy and map-based trajectory error ($E^{map}_{trajectory}$, Eq.~(\ref{eq.map_traj})) for direct localization precision.}

\vspace{-0.7cm}
\textcolor{black}{\subsubsection{Multi-Map Configuration}}
\textcolor{black}{We evaluate multi-map fusion using two scenarios: non-overlapping maps on NC dataset and overlapping maps on ZJG dataset. Fig.\ref{fig:multimap_align} shows the spatial distribution of observations from each map. For each scenario, we select two sub-maps and compare individual versus fused localization results.}

\textcolor{black}{Table~\ref{tab:multi-map-compare} shows that multi-map fusion consistently reduces both mean error and standard deviation. On ZJG-qsdjt-1112, fusion achieves 1.85m error versus 1.98m and 2.19m for individual maps. On NC-quad-hard, error decreases from 0.22m and 0.17m (single maps) to 0.16m (fused). The standard deviation reduction indicates more stable estimation.}

\textcolor{black}{To assess extrinsic estimation accuracy, we transform both maps to the VIO frame using estimated extrinsic parameters. Fig.\ref{fig:multimap_align} visualizes the alignment, where brown and green represent Map 1 and Map 2. The aligned trajectories show minimal divergence, and map edge contours remain well-defined, suggesting accurate extrinsic estimation.}

\begin{table}[tp]
\caption{The translation part (m) of local-based trajectory error under different number of maps.}\label{tab:multi-map-compare}
    \centering
\resizebox{0.5\textwidth}{!}{
\begin{tabular}{lcccccc}
\hline \hline
      \multirow{2}{*}{Scene} & \multicolumn{2}{c}{map1} & \multicolumn{2}{c}{map2} & \multicolumn{2}{c}{2 maps} \\ 
      & mean        & std        & mean        & std        & mean         & std         \\ \hline
ZJG-yyl-0304   & 2.24        & 0.68       & 3.50        & 2.26       & 2.23         & 0.63        \\
ZJG-qsdjt-1112 & 1.98        & 1.26       & 2.19        & 1.14       & 1.85         & 1.06        \\
NC-math-hard  & 1.22        & 0.42       & 0.44        & 0.23       & 0.47         & 0.19        \\
NC-quad-hard  & 0.22        & 0.18       & 0.17        & 0.09       & 0.16         & 0.07   \\  \hline\hline  
\end{tabular}}
\vspace{-0.2cm}
\end{table}

\begin{table}[tp]
     \caption{The translation part (m) of map-based trajectory error under different numbers of camera observations used for localization.}\label{table:multicam-loc}
    \centering
    \resizebox{0.425\textwidth}{!}{
    \begin{tabular}{lcccc}
    \hline\hline
        ~ & 1 cam & 2 cam & 3 cam & 4 cam \\ \hline
        ZJG-qsdjt-1112 & 6.77 & 6.76 & 4.85 & 4.28 \\ 
        ZJG-yyl-0304 & 6.5 & 6.45 & 4.18 & 3.62 \\ 
        NC-math-hard & 0.56 & 0.54 & 0.64 & 0.66 \\ 
        NC-quad-hard & 0.44 & 0.49 & 0.45 & 0.26 \\
        Average & 3.57& 3.56& 2.53& 2.21\\\hline\hline
    \end{tabular}}
\vspace{-0.2cm} 
\end{table}

\begin{table}[tp]
  \centering
  \caption{Comparison of successful loop closure detections by monocular and multi-camera algorithms under varying feature counts.}
  \label{table:loop_compare}
  \scalebox{0.88}{ 
  \begin{threeparttable}
  \setlength{\tabcolsep}{4pt}
  \renewcommand{\arraystretch}{1.1}
  \begin{tabular}{@{}>{\centering\arraybackslash}m{2cm} c *{3}{w{c}{1.2cm}} @{}}
  \hline\hline
  \multirow{2}*{\centering Dataset} & \multirow{2}*{Method} & \multicolumn{3}{c}{Number of feature points} \\
  \cline{3-5}
  & & 50 & 100 & 150 \\ 
  \hline
  \multirow{2}*{\centering ZJG} 
    & Mono\textsuperscript{1}  & 113 & 554 & 706 \\
    & MulCam\textsuperscript{2} & 498 & 878 & 954 \\ 
  \hline
  \multirow{2}*{\centering NC} 
    & Mono\textsuperscript{1}  & 6   & 71  & 140 \\
    & MulCam\textsuperscript{2} & 151 & 186 & 213 \\
  \hline\hline
  \end{tabular}
  \begin{tablenotes}
    \footnotesize
    \item[1] Mono: Single forward-facing camera
    \item[2] MulCam: Multiple cameras (forward + surround)
  \end{tablenotes}
  \end{threeparttable}
  \vspace{-0.2cm}
  }
\end{table}

\begin{figure}[tp]
    \centering
    \includegraphics[width=0.5\textwidth]{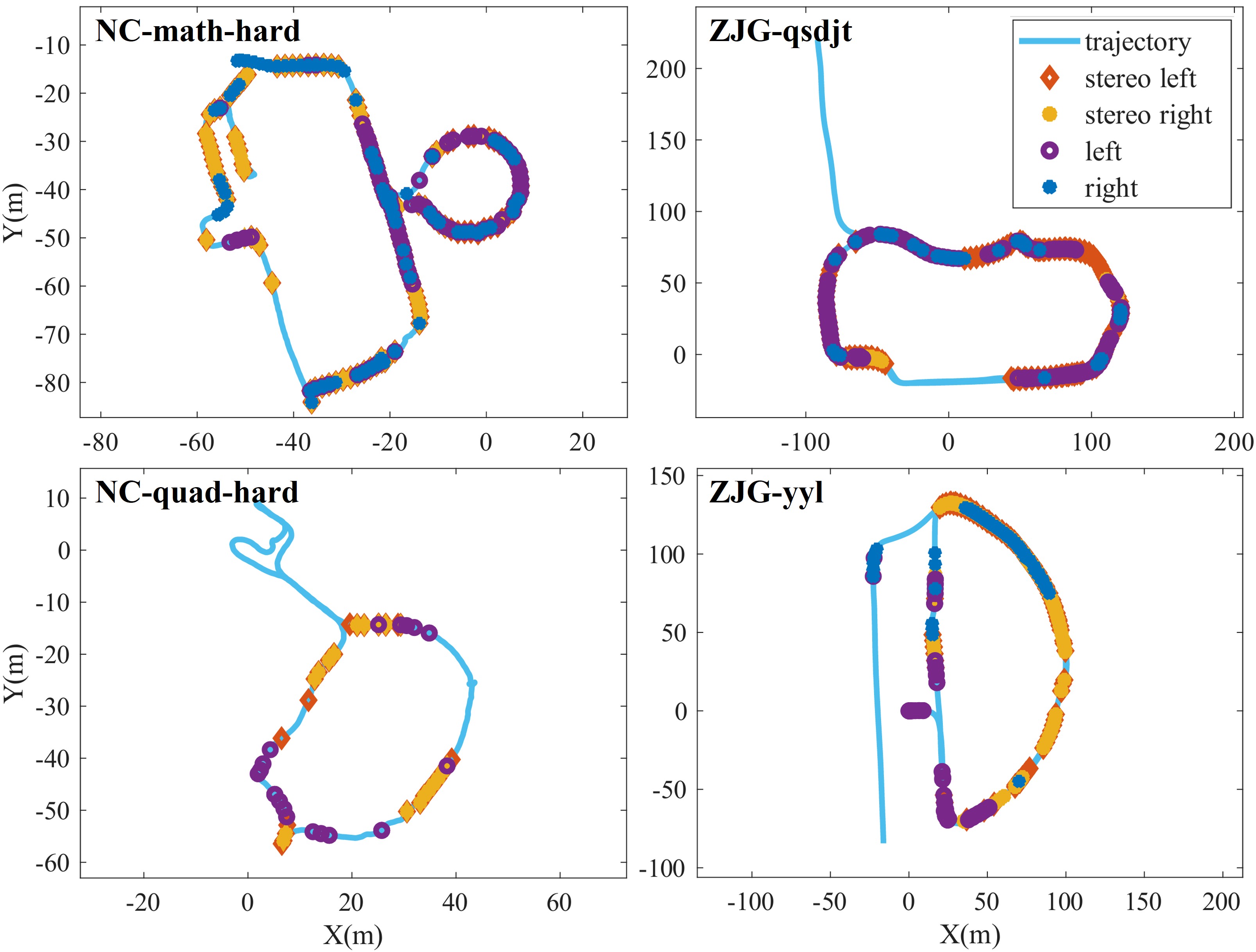}
    \caption{Visualization results of localization observations (points with at least 40 feature matches) from different camera types on different datasets. Key: blue diamonds (stereo left), red stars (stereo right), green circles (left surround), magenta stars (right surround).}
    \label{fig:loc_visualization}
    \vspace{-0.4cm}
\end{figure}

\vspace{-0.2cm}
\textcolor{black}{\subsubsection{Multi-Camera Configuration}}
\textcolor{black}{We test localization accuracy using 1 to 4 cameras from our multi-camera platform (Fig.\ref{fig:car_setup}, Table~\ref{tab:sensor_config}). The configurations are: '1cam' (cam2 only), '2cam' (cam2+cam3), '3cam' (cam2+cam3+cam6), and '4cam' (cam2+cam3+cam4+cam6).}

\textcolor{black}{Table~\ref{table:multicam-loc} demonstrates that multi-camera fusion reduces average error from 3.57m with a single camera to 2.21m with four cameras, with gains varying by environment.}

\textcolor{black}{Fig.\ref{fig:loc_visualization} explains this pattern by showing observation distributions. On ZJG, surround cameras capture significantly more valid features than front cameras, justifying the large accuracy gains. On NC, observations distribute more evenly across cameras, correlating with smaller improvements.}

\begin{figure}[tp]
    \centering
\includegraphics[width=0.48\textwidth]{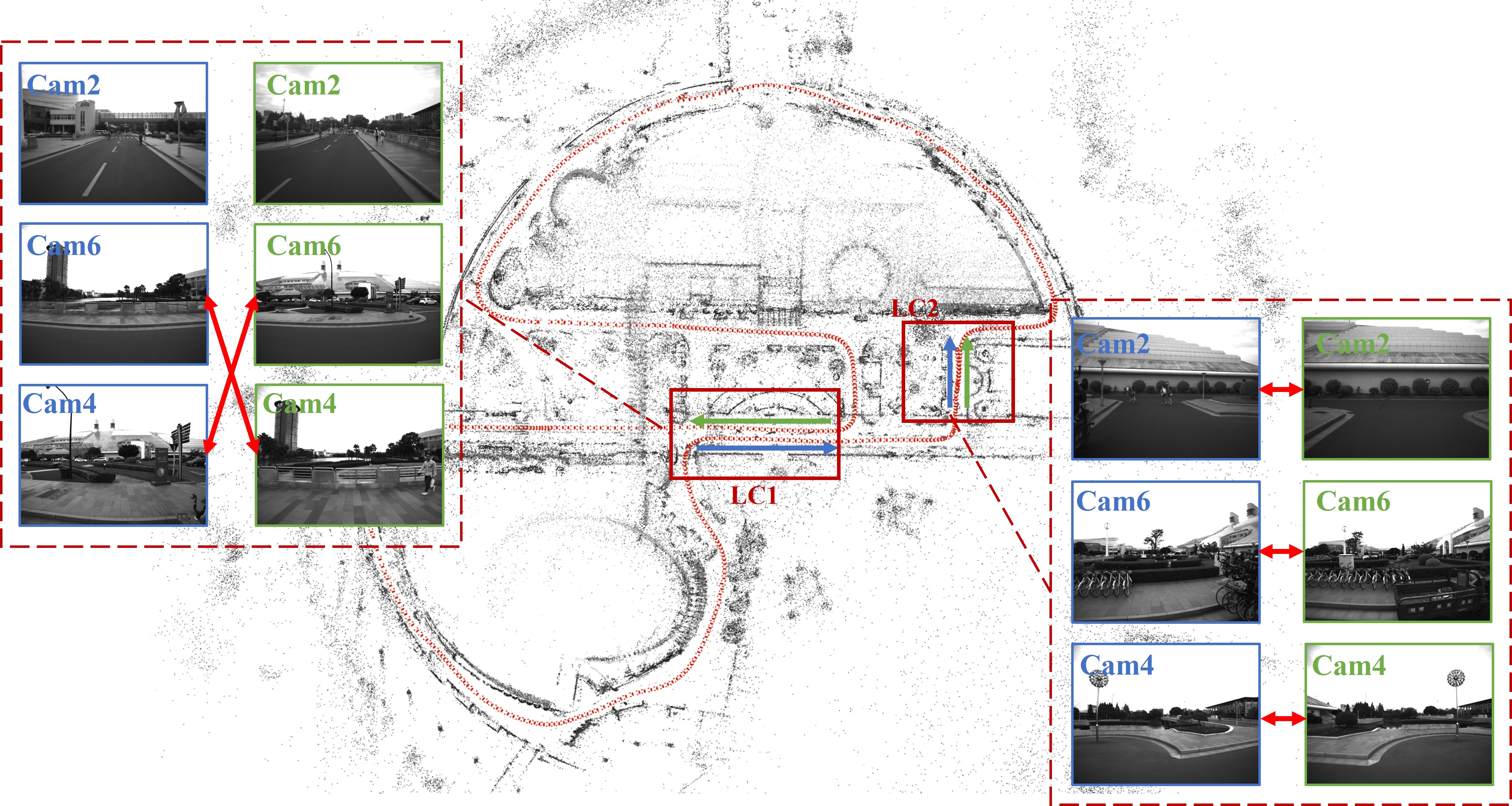}
    \caption{Visualization results of loop-closure observations in the ZJG dataset. Blue and green arrows indicate robot poses at loop closure moments (LC1, LC2). Image insets show views from different cameras (Cam2: stereo left, Cam6/4: surround views). Multi-camera fusion enables loop closure (LC1) even with opposing viewpoints where the monocular front camera fails.}
    \label{fig:loop_presentation}
\vspace{-0.2cm}
\end{figure}

\textcolor{black}{Beyond localization, multi-camera observations also enhance mapping robustness through improved loop closure detection. To validate this, we reduce detectable feature points per image to 50, 100, and 150, then compare successfully detected loop closures using monocular versus multi-camera configurations on the same datasets. Table~\ref{table:loop_compare} shows that monocular loop closure degrades severely under feature-sparse conditions: on NC dataset with 50 features, monocular detects only 6 loops while multi-camera achieves 151. On ZJG, multi-camera maintains 498 detections versus 113 for monocular.}
\textcolor{black}{The advantage of multi-camera LCD is further illustrated qualitatively in Fig.\ref{fig:loop_presentation}. In scenario LC2, where robot orientations are similar, both mono and multi-camera methods might detect the loop. However, in challenging cases like LC1 with opposing viewpoints, the front-facing camera (cam2) often fails due to the large visual difference, while the surround cameras (cam6/cam4) capture consistent structural information, enabling successful loop closure by the multi-camera system. This confirms the necessity of leveraging multiple viewpoints for robust loop detection, as implemented in our system.}

\vspace{-0.2cm}
\textcolor{black}{\subsubsection{Discussion and Practical Implications}}
\textcolor{black}{The experiments demonstrate that multi-map and multi-camera configurations both enhance localization robustness, though through distinct mechanisms. Multi-map fusion reduces error by 15-60\% across scenarios (Table~\ref{tab:multi-map-compare}), providing redundancy when individual maps contain poorly reconstructed regions or occlusions. The system dynamically weights observations based on uncertainty, effectively rejecting outlier constraints while maintaining computational efficiency through filter-based architecture.}

\textcolor{black}{Multi-camera configurations provide dual benefits for both mapping and localization. During mapping, surround cameras maintain loop closure detection under feature-sparse conditions where monocular approaches fail (Table~\ref{table:loop_compare}), ensuring reconstruction quality. During localization, environment-dependent gains emerge: feature-sparse ZJG sequences achieve 40\% error reduction when expanding from 2 to 4 cameras, as lateral views capture structural information absent in forward perspectives (Fig.\ref{fig:loc_visualization}). Feature-rich NC sequences show modest improvements, indicating the system adapts to available observability without over-reliance on sensor redundancy. For deployment, this architecture provides systematic robustness across the full pipeline. Large outdoor environments benefit from complete multi-camera coverage throughout mapping and localization phases, while compact spaces maintain reliable operation with reduced configurations, preserving versatility without compromising accuracy.}

\vspace{0.6cm}
\subsection{REAL-TIME SYSTEM EVALUATION}\label{sec:realtime}
\textcolor{black}{The previous sections validate individual components, including mapping quality, matching robustness, and localization accuracy. This section evaluates the complete system's real-time performance in practical deployment scenarios where robots operate in previously mapped environments. We compare our approach against state-of-the-art visual-inertial SLAM systems (ORB-SLAM3, VINS-Fusion, Maplab) that support map-based localization, focusing on causal trajectory accuracy suitable for robot control.}

\textcolor{black}{Two complementary evaluation scenarios assess system performance. First, we measure local-based trajectory error ($E^{local}_{trajectory}$, Eq.~(\ref{eq.local_traj})) on NC and ZJG datasets to evaluate map-constrained tracking accuracy after initialization, which directly reflects control-loop feedback quality. Second, we measure map-based trajectory error ($E^{map}_{trajectory}$, Eq.~(\ref{eq.map_traj})) on EuRoC dataset to assess end-to-end localization precision in controlled indoor environments. For fair comparison, all methods operate without non-causal loop closure optimization during localization, and each builds maps using its native pipeline. Our method constructs maps using FAST-LIO2 external poses on NC/ZJG, and pure visual-inertial mapping on EuRoC.}

\vspace{-0.2cm}
\subsubsection{Outdoor Large-Scale Evaluation}
\textcolor{black}{We evaluate local-based trajectory accuracy on NC and ZJG datasets, which present diverse challenges including seasonal variations, illumination changes, and scale differences. We compare against ORB-SLAM3 and VINS-Fusion; Maplab is excluded as its front-end (ROVIO) fails to initialize reliably in most outdoor scenarios. For ORB-SLAM3, we record post-localization trajectories to avoid non-causal pose discontinuities. Table~\ref{all-ate-compare} presents results across eight scenarios: four from NC (naming: location-difficulty) and four from ZJG (naming: location-time).}

\textcolor{black}{Our method achieves the lowest average error (4.72m) and standard deviation (1.82m) across all scenarios, with no tracking failures. VINS-Fusion exhibits tracking failures on two challenging sequences (math-hard, yyl-1022) but performs competitively on feature-rich scenarios (quad-hard: 1.06m). ORB-SLAM3 shows severe drift on math-hard (34.05m versus our 3.94m) and fails on quad-hard and qsdjt-0224. Notably, on the most challenging sequence (qsdjt-0224), our method maintains 15.17m error while VINS-Fusion drifts to 24.74m and ORB-SLAM3 fails completely.}

\begin{table}[t]\small
    \caption{The translation part (m) of local-based trajectory accuracy of each method operating on various challenge environments.}\label{all-ate-compare}
    \centering
\begin{threeparttable}
\begin{tabular}{lcccccc}
\hline \hline
                   & \multicolumn{2}{c}{VILO(ours)} & \multicolumn{2}{c}{VINS-Fusion}                               & \multicolumn{2}{c}{ORB-SLAM3}                               \\
\multirow{-2}{*}{} & mean           & std          & mean                   & std                           & mean              & std                         \\ \hline
quad-easy          & \textbf{1.24}  & 0.59 & 1.81                          & \textbf{0.47}                          & 2.97 & 1.22 \\
quad-hard          & 1.58           & \textbf{0.79} & \textbf{1.06}                 & 0.85                          & fail                        & fail                        \\
math-hard          & \textbf{3.94}  & \textbf{1.76} & fail                          & fail                          & 34.05                        & 14.76                        \\
math-medium        & \textbf{1.18}  & \textbf{0.60} & 1.58                          & 0.66                          & 13.59                        & 6.11                         \\
yyl-1022            & \textbf{2.78}  & \textbf{1.58}          & fail                          & fail                          & 7.61 & 3.67 \\
yyl-1112            & \textbf{4.23}           & 2.08 & 4.84  & 2.19  & 4.29 & \textbf{1.80}\\
qsdjt-1022          & \textbf{7.67}  & \textbf{3.41} & 14.03                         & 6.56                          & 4.55                         & 3.08                         \\
qsdjt-0224          & \textbf{15.17}           & \textbf{3.71} & 24.74 & 25.06 & fail                         & fail    \\ \hline  
Average & \textbf{4.72}& \textbf{1.82} & 8.01 & 5.97  & 11.18  &  5.11  \\
\hline \hline
\end{tabular}
\begin{tablenotes}
\item[1] fail indicates a great drift in the odometer during the movement.
\end{tablenotes}
\end{threeparttable}
\vspace{-0.2cm}
\end{table}

\begin{table}[t]\small
    \caption{The translation part (m) of the map-based trajectory error of each method operating on EuRoC dataset.}\label{all-ate-compare-EuRoC}
    \centering
    \setlength{\tabcolsep}{1.5mm}{
\begin{threeparttable}
\begin{tabular}{ccccccccc}
\hline \hline
     & \multicolumn{2}{l}{VILO(ours)} & \multicolumn{2}{l}{ORB-SLAM3} & \multicolumn{2}{l}{VINS-Fusion} & \multicolumn{2}{l}{Maplab} \\  
     & mean        & std        & mean          & std           & mean           & std            & mean         & std         \\ \hline
MH02 & \textbf{0.18}        & \textbf{0.08}       & 1.82          & 2.25          & 0.38           & 0.11           & 1.45         & 0.61        \\
MH03 & \textbf{0.29}        & \textbf{0.09}       & 0.21          & 0.38          & 0.36           & 0.14           & 0.92         & 0.39        \\
MH04 & \textbf{0.43}        & 0.16       & fail          & fail          & 0.53           & \textbf{0.11}           & 1.78         & 0.58        \\
MH05 & 0.59        & 0.34        & fail          & fail         & \textbf{0.56}           & \textbf{0.17}           & 2.06         & 0.74        \\
V102 & 0.18        & 0.05       & 0.49          & 0.89          & 0.19           & 0.04           & \textbf{0.06}         & \textbf{0.04}        \\
V103 & 0.24        & 0.08    & 0.27          & 0.69          & 2.82           & 0.72           & \textbf{0.10}         & \textbf{0.04}        \\
V202 & 0.12        & \textbf{0.05}       & 0.32          & 0.79          & 0.73            & 0.40           & \textbf{0.08}        & 0.07       \\
V203 & 0.25        & 0.1       & fail\footnotemark[1]          & fail          & 1.03           & 0.59           & \textbf{0.21}          & \textbf{0.09}       \\ \hline
Average & \textbf{0.29} & \textbf{0.12} & 0.62 & 1.0 & 0.83 & 0.29 & 0.83 & 0.32 \\
\hline \hline
\end{tabular}
\begin{tablenotes}
\item[1] fail indicates there is no successful localization during the movement.
\end{tablenotes}
\end{threeparttable}}
\vspace{-0.2cm}
\end{table}

\vspace{-0.2cm}
\subsubsection{Indoor Controlled Evaluation}
\textcolor{black}{We evaluate map-based trajectory accuracy on EuRoC dataset~\cite{EuRoC}, which provides controlled indoor environments with motion capture ground truth. We construct maps using MH01, V101, and V201 sessions with each method's native mapping pipeline (our method uses pure visual-inertial mapping without external pose data), then perform localization on remaining sequences in each environment group. To measure map-based accuracy, we align each method's mapping trajectory with ground truth via least squares, then transform the ground truth localization trajectories into the constructed map coordinate system using the obtained transformation matrix. Table~\ref{all-ate-compare-EuRoC} presents results.}

\textcolor{black}{Our method achieves the best average accuracy (0.29m) and lowest standard deviation (0.12m) across all eight sequences. Maplab consistently outperforms all methods in Vicon Room scenarios (V102/V103/V202/V203: 0.06-0.21m), leveraging its batch optimization approach to fully exploit dense feature constraints in small-scale, structured environments. However, this specialization comes at the cost of degraded performance in larger Machine Hall scenarios (MH02-MH05: 0.92-2.06m versus our 0.18-0.59m), where the optimization-based approach struggles with scale variations. ORB-SLAM3 fails on three sequences (MH04, MH05, V203), while VINS-Fusion exhibits both lower accuracy (0.83m average) and higher variance (0.29m std) compared to our system.}

\vspace{-0.4cm}
\textcolor{black}{\subsubsection{Discussion and Practical Implications}}
\textcolor{black}{The experiments reveal fundamental trade-offs between filter-based and optimization-based localization architectures across operational scales. In large-scale outdoor environments, the proposed method achieves superior robustness (zero failures versus four total failures across baselines) through three design elements: deterministic initialization ensures reliable startup under extreme outlier rates, IMU-aided minimal solvers maintain feature matching under seasonal variations, and multi-map fusion provides redundancy against individual map deficiencies. This advantage is most pronounced in challenging scenarios where our method maintains bounded error while ORB-SLAM3 exhibits order-of-magnitude drift or complete failure. However, in small-scale controlled environments, batch optimization methods like Maplab achieve superior accuracy by fully exploiting dense feature constraints, though at the cost of degraded performance when environment scale increases.}

\textcolor{black}{The critical distinction lies in cross-environment consistency rather than peak performance. Our filter-based architecture demonstrates balanced accuracy across diverse conditions, remaining competitive in compact spaces while maintaining precision in large environments, all without requiring environment-specific tuning. For field robotics applications, this operational reliability is paramount: guaranteed localization across unpredictable conditions justifies modest accuracy trade-offs in idealized scenarios. Systems requiring assured operation under long-term appearance changes, illumination variations, and scale diversity should prioritize the proposed architecture. While controlled indoor deployments with stable conditions may show smaller performance gaps compared to non-causal optimization methods in offline analysis, the causality requirement remains essential for real-time robot control regardless of environment type.}

\textcolor{black}{\subsection{Summary of Experimental Findings}}
\textcolor{black}{The comprehensive evaluation across mapping, matching, localization, and integrated system performance reveals systematic design principles for field deployment. In the mapping stage, loop closure-enhanced VINS priors substantially improve reconstruction coherence, while soft extrinsic constraints prevent calibration errors from degrading multi-camera map quality. External pose sensors provide measurable benefits primarily for trajectories exceeding 500 meters, where visual-inertial drift begins to dominate the error budget. During initialization and online matching, the deterministic approach achieves zero-variance convergence under outlier rates up to 62\%, while IMU-aided 2-point minimal solvers demonstrate 2-3$\times$ higher success rates than traditional 3-4 point methods in challenging outdoor scenarios. For real-time localization, multi-map fusion consistently reduces trajectory error through constraint redundancy, with improvements ranging from 16\% to 61\% depending on individual map quality. Multi-camera configurations show environment-dependent benefits, yielding about 40\% error reduction in feature-sparse outdoor scenes (ZJG sequences), while offering limited gains in structured environments. System-level comparisons demonstrate zero tracking failures across all test scenarios, maintaining 4.72m average error compared to 8.01m (VINS-Fusion) and 11.18m (ORB-SLAM3) in large-scale outdoor sequences where baseline methods exhibit severe drift or complete failure.}

\textcolor{black}{These findings also reveal operational limitations. Localization accuracy fundamentally depends on offline map quality, as reconstruction errors propagate directly to online estimation. Feature-sparse environments challenge both loop closure detection during mapping and matching success during localization, though multi-camera configurations substantially mitigate these effects. The deterministic initialization incurs 0.1-0.7 seconds of computational overhead, acceptable for system startup but prohibitive for continuous re-localization scenarios requiring real-time performance.}

\vspace{0.4cm}
\section{CONCLUSION}\label{sec:conclusion}
\textcolor{black}{We present a visual-inertial localization system providing causal, map-constrained pose estimates for robot control loops. Our contributions address four fundamental challenges: (a) a multi-camera architecture improving both mapping robustness and localization accuracy through expanded observability, (b) a multi-map fusion framework enabling operation across disconnected regions without overlap requirements, (c) robust matching under extreme outlier rates exceeding 80\% through deterministic initialization and IMU-aided 2-point minimal solvers, and (d) a comprehensive evaluation framework measuring causality-preserving performance aligned with control requirements.}

\textcolor{black}{Comprehensive validation across diverse indoor and outdoor scenarios confirms the system's robust operation under varying illumination, seasonal changes, ongoing construction, and different environmental scales. Through these experiments, we establish systematic design principles for deploying multi-camera configurations, multi-map fusion, and external sensor integration based on environment characteristics and trajectory scale.}


\textcolor{black}{The open-source release of both system and dataset aims to facilitate research bridging the gap between SLAM accuracy and control-oriented reliability in field robotics.}


\ifCLASSOPTIONcaptionsoff
  \newpage
\fi

\bibliographystyle{IEEEtran}
\bibliography{root}

\end{document}